\documentclass[10pt,journal,compsoc]{IEEEtran}

\ifCLASSOPTIONcompsoc

  \usepackage[nocompress]{cite}
\else

  \usepackage{cite}
\fi

\ifCLASSINFOpdf
 
\else
 
\fi

\usepackage{svg} % 需使用包
\usepackage{amsmath}
\usepackage{subfig}
\usepackage{amssymb}
\usepackage{enumitem}
\usepackage{algorithm}
\usepackage{multirow}

\usepackage{mathrsfs}
\usepackage{amssymb}
\usepackage{bm}
\usepackage{dsfont}
\usepackage{multirow}
\usepackage{comment}
\usepackage{amsthm}
\theoremstyle{plain}
\theoremstyle{definition}
\usepackage[normalem]{ulem}
\usepackage{makecell}

\usepackage{tikz}
\usepackage{times}
\usepackage{dsfont}
\usepackage{subfig}
\usepackage{helvet}
\usepackage{subfig}
\usepackage{amssymb}
\usepackage{courier}
\usepackage{ragged2e}
\usepackage{multirow}
\usepackage{mathrsfs}
\usepackage{graphicx}
\usepackage{blindtext}
\usepackage{algorithmic}
\usepackage{lineno,hyperref}
\usepackage[marginal]{footmisc}
\usepackage{bbm}
\usepackage[utf8]{inputenc}
\usepackage[english]{babel}
\usepackage{epstopdf}
\usepackage{array}
\usepackage{booktabs}

\begin{document}
% ,~Xifeng~Guo,~Siwei~Wang
\title{A Survey of Deep Graph Clustering: Taxonomy, \\ Challenge, Application, and Open Resource}
% ~Shirui~Pan,~\IEEEmembership{Senior~Member,~IEEE}
% ~Cheng~Tan,
\author{Yue~Liu,~Jun~Xia,~Sihang~Zhou,\\~Xihong~Yang,~Ke~Liang,~Chenchen~Fan,~Yan~Zhuang,\\~Stan~Z.~Li,~\IEEEmembership{Fellow,~IEEE},~Xinwang~Liu$^{\dagger}$,~\IEEEmembership{Senior~Member,~IEEE},~Kunlun~He$^{\dagger}$
\IEEEcompsocitemizethanks{
\IEEEcompsocthanksitem Y. Liu, S. Zhou, X. Yang, K. Liang, and X. Liu are with National University of Defense Technology, Changsha, China. (E-mail: yueliu19990731@163.com). J. Xia and Stan Z. Li are with Westlake University, Hangzhou, China. C. Fan and Y. Zhuang are with Big Data Research Center, Chinese PLA General Hospital, Beijing, China. 

\IEEEcompsocthanksitem $^{\dagger}$: Corresponding author. 

% \IEEEcompsocthanksitem Manuscript was received on Sept. 3, 2023.
}
}

% \markboth{arXiv pre-print}{Y. Liu \MakeLowercase{\textit{et al.}}: A Survey of Deep Graph Clustering: Taxonomy, Challenge, Application, and Open Resource}

\IEEEtitleabstractindextext{%
\justifying
\begin{abstract}
Graph clustering, which aims to divide nodes in the graph into several distinct clusters, is a fundamental yet challenging task. Benefiting from the powerful representation capability of deep learning, deep graph clustering methods have achieved great success in recent years. However, the corresponding survey paper is relatively scarce, and it is imminent to make a summary of this field. From this motivation, we conduct a comprehensive survey of deep graph clustering. Firstly, we introduce formulaic definition, evaluation, and development in this field. Secondly, the taxonomy of deep graph clustering methods is presented based on four different criteria, including graph type, network architecture, learning paradigm, and clustering method. Thirdly, we carefully analyze the existing methods via extensive experiments and summarize the challenges and opportunities from five perspectives, including graph data quality, stability, scalability, discriminative capability, and unknown cluster number. Besides, the applications of deep graph clustering methods in six domains, including computer vision, natural language processing, recommendation systems, social network analyses, bioinformatics, and medical science, are presented. Last but not least, this paper provides open resource supports, including 1) a collection (\url{https://github.com/yueliu1999/Awesome-Deep-Graph-Clustering}) of state-of-the-art deep graph clustering methods (papers, codes, and datasets) and 2) a unified framework (\url{https://github.com/Marigoldwu/A-Unified-Framework-for-Deep-Attribute-Graph-Clustering}) of deep graph clustering. We hope this work can serve as a quick guide and help researchers overcome challenges in this vibrant field. 
\end{abstract}

}

\maketitle

\IEEEdisplaynontitleabstractindextext

\IEEEpeerreviewmaketitle

\IEEEraisesectionheading{\section{Introduction}}

\IEEEPARstart{G}raph clustering is an important and challenging task to separate nodes of the graph into different clusters in an unsupervised manner. In recent years, benefiting from the strong representation capability of deep learning \cite{deep_learning}, especially graph neural networks (GNNs) \cite{GCN,GAT,GAE,MGCN,LIANGKE1,LIANGKE2,LIANGKE3}, deep graph clustering has witnessed fruitful advances. However, unlike the deep clustering area \cite{deep_clustering_survey_1,deep_clustering_survey_2,deep_clustering_survey_3,deep_clustering_survey_4}, the survey paper of deep graph clustering \cite{survey_1} is relatively scarce. For example, \cite{survey_2,survey_3} mainly survey the papers about community detection with deep learning. To better assist researchers in reviewing, summarizing, and planning for the future, a comprehensive survey of deep graph clustering is expected. From this motivation, we make a comprehensive survey of deep graph clustering in this paper.

\begin{figure}[h]
\centering
\small
\begin{minipage}{0.9\linewidth}
\centerline{\includegraphics[width=1\textwidth]{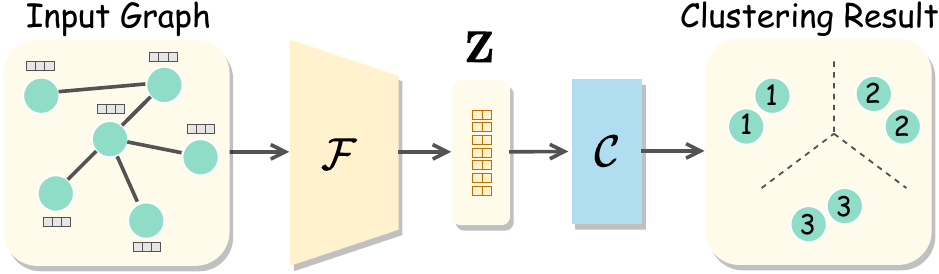}}
\end{minipage}
\caption{The general pipeline of deep graph clustering.}
\label{overall}
\end{figure}

Firstly, we demonstrate the general pipeline of deep graph clustering. As shown in Figure \ref{overall}, the encoding neural network $\mathcal{F}$ is trained in a self-supervised manner to embed the nodes into the latent space. After encoding, the clustering method $\mathcal{C}$ separates the node embeddings $\mathbf{Z}$ into several disjoint clusters. The formulaic definitions and the important baselines of deep graph clustering are deliberated in Section \ref{sec:deep_graph_clusering}.

% , which are trained in an unsupervised manner, 

\begin{figure}[h]
\centering
\small
\begin{minipage}{0.9\linewidth}
\centerline{\includegraphics[width=1\textwidth]{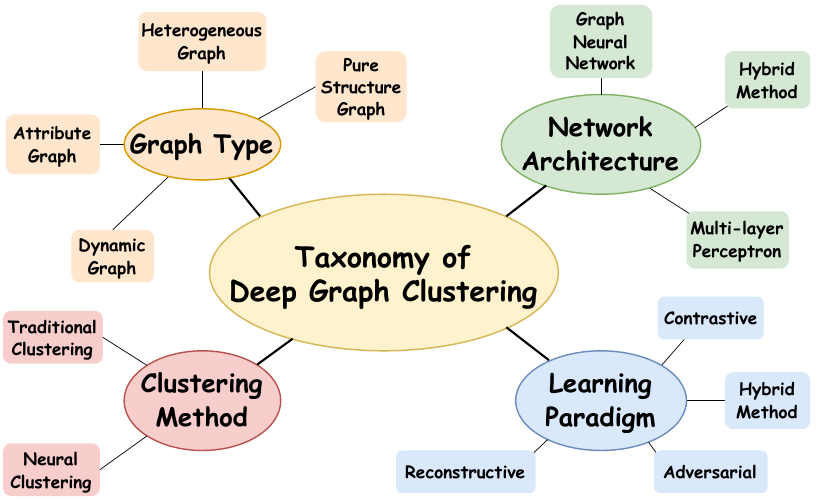}}
\end{minipage}
\caption{The taxonomy of deep graph clustering.}
\label{taxonomy_figure}
\end{figure}

Secondly, as shown in Figure \ref{taxonomy_figure}, we contribute a structured taxonomy to provide a broad overview of this field, categorizing existing works from four perspectives: graph type, network architecture, learning paradigm, and clustering method. More specifically, the input graph type can be classified into four distinct categories: pure structure graph, attribute graph, heterogeneous graph, and dynamic graph. We analyze the characteristics of each graph type and introduce the corresponding processing methods. Besides, for the network architecture, the existing deep graph clustering methods are grouped into multi-layer perceptron-based (MLP-based) methods, graph-neural network-based (GNN-based) methods, and hybrid methods. The benefits and drawbacks of each type are carefully discussed. Moreover, the learning paradigms are classified into reconstructive, adversarial, contrastive, and hybrid paradigms. For each learning paradigm, the general pipeline is summarized in detail. In addition, the clustering methods are separated into traditional clustering methods and neural clustering methods. The advantages and disadvantages of traditional clustering are analyzed, and the technological evolution of neural clustering is summarized in depth. We elaborate on the taxonomy of deep graph clustering in Section \ref{taxonomy}.

% and analyze the existing methods

\begin{figure}[h]
\centering
\small
\begin{minipage}{0.9\linewidth}
\centerline{\includegraphics[width=1\textwidth]{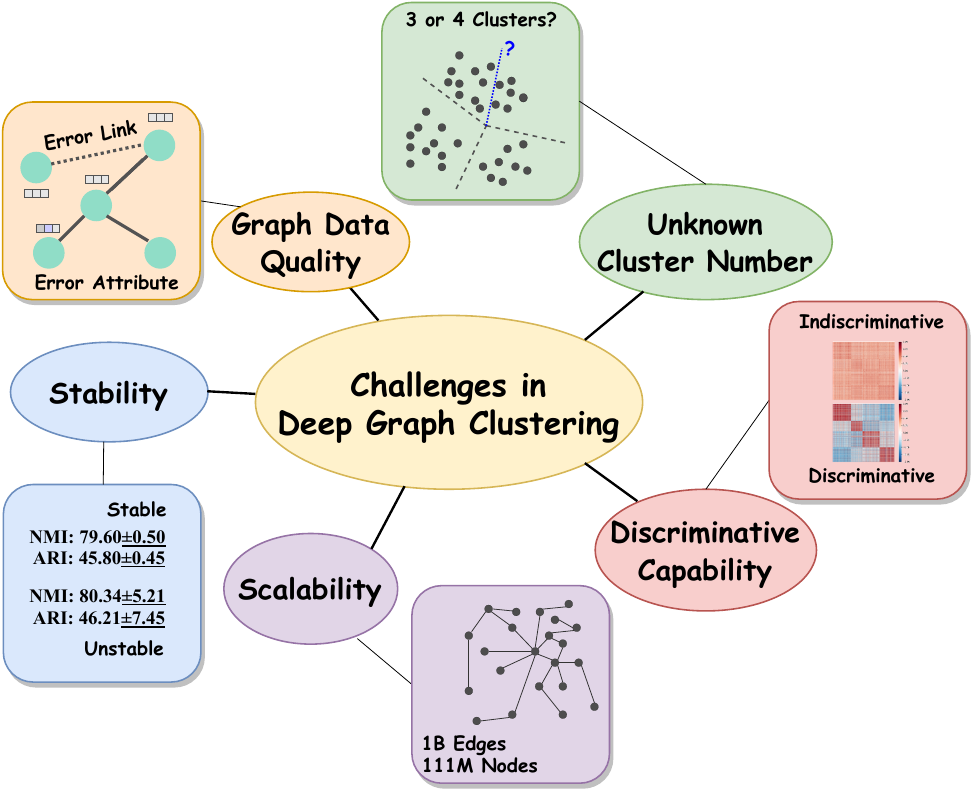}}
\end{minipage}
\caption{The challenges in deep graph clustering.}
\label{challenges_figure}
\end{figure}

Despite the remarkable progress, this fast-growing field is still fraught with several crucial challenges. Therefore, we conduct comprehensive experiments to summarize and analyze the challenges of deep graph clustering. In addition, we carefully discuss the potential opportunities to solve the challenges in this field. Specifically, Figure \ref{challenges_figure} demonstrates the problems of graph data quality, stability, scalability, discriminative capability, and unknown cluster number. The detailed analysis and the potential solutions are provided in Section \ref{challenge_opportunity}.

% Therefore, we conduct comprehensive experiments to analyze the challenges of deep graph clustering. 

% , such as, graph data quality, stability, scalability, discriminative capability, and unknown cluster number. 

 % We hope this work will become a quick guide to deep graph clustering and motivate the researchers to overcome the challenges in this area. 

Moreover, in recent years, deep graph clustering methods have been successfully applied to numerous domains, such as social network analysis, recommendation systems, computer vision, natural language processing, bioinformatics, medical science, etc. The detailed information refers to Section \ref{application_section}. The main contributions of this paper are summarized as follows.

\begin{itemize}
\item We present a comprehensive survey paper in deep graph clustering domain to help researchers review, summarize, solve challenges, and plan for future.

% In order to help the researchers review, summarize, and plan for future, we present a comprehensive survey paper in deep graph clustering domain. 

% the comprehensive survey paper on deep graph clustering is presented. 

\item We design a taxonomy of recent deep graph clustering methods based on four aspects, i.e., graph type, network architecture, learning paradigm, and clustering method.

\item We conduct extensive experiments to summarize the challenges in the deep graph clustering field from five perspectives, including graph data quality, stability, scalability, discriminative capability, and unknown cluster number. Besides, through the careful analyses, we provide the potential promising technical solutions.

\item We share two practical open resources, including a collection of state-of-the-art deep graph clustering methods and a unified framework of deep graph clustering.

\end{itemize}

\section{Deep Graph Clustering} \label{sec:deep_graph_clusering}
In this section, we first introduce the basic notation and the formulaic definition of deep graph clustering. Then the critical deep graph clustering baselines are deliberated.

\subsection{Notation}\label{define}
The basic notations in this paper are summarized in Table \ref{NOTATION_TABLE}. First, we take the attribute graph as an example input of the deep graph clustering methods. Given an attribute graph $\mathcal{G}_{\text{A}} = \{\mathcal{V}, \mathcal{E}, \mathbf{X}\}$, $\mathcal{V}=\{v_1, v_2, \dots, v_n\}$ denotes a node set, which contains $k$ classes of $n$ nodes. Besides, $\mathcal{E}$ denotes a set of edges that link nodes. In addition, each node in the graph attaches $d'$-dimension attributes. In matrix form, $\mathbf{X} \in \mathbb{R}^{n\times d'}$ denotes the node attribute matrix and $\mathbf{A}\in \mathbb{R}^{n \times n}$ denotes the adjacency matrix. $\mathbf{y} \in \mathbb{R}^{n}$ denotes ground truth labels of nodes. We introduce other types of the input graph in Section \ref{graph_type_section}.

\subsection{Task Definition}

% For other types of the input graph, please check in Section . 

% and $\mathbf{X} \in \mathbb{R}^{N\times D}$ is the node attribute matrix. 

% Also, with matrix form, $\mathcal{G}_{\text{A}}$ is represented by the adjacency matrix $\mathbf{A}\in \mathbb{R}^{N \times N}$ and the attribute matrix $\mathbf{X} \in \mathbb{R}^{N\times D}$.

% As shown in Figure \ref{overall}, the target of deep graph clustering is to encode nodes with neural networks and divide them into different clusters. 

\begin{table}[!t]
\centering
% \small
\caption{Basic notation summary.}
\resizebox{\linewidth}{!}{
% \scalebox{1.1}{
\begin{tabular}{ll}
\toprule
\textbf{Notations}                                        & \textbf{Meanings}                                \\ \midrule
$\mathcal{G}_{\text{S}}$  & Pure structure graph  
\\
$\mathcal{G}_{\text{A}}$  & Attribute graph  
\\
$\mathcal{G}_{\text{H}}$  & Heterogeneous graph  
\\
$\mathcal{G}_{\text{D}}$  & Dynamic graph  
\\
$\mathcal{V}$  & Set of nodes
\\
$\mathcal{E}$  & Set of edges  
\\
$\mathcal{F}$   & Encoding network 
\\
$\mathcal{C}$   & Clustering method 
\\
$\mathcal{M}$   & Evaluation method
\\
$n \in \mathbb{R}$   & Sample number 
\\
$d' \in \mathbb{R}$   & Attribute number 
\\
$d \in \mathbb{R}$   & Dimension number of learned feature 
\\
$k \in \mathbb{R}$   & Cluster number 
\\
$s \in \mathbb{R}$   & Clustering score
\\
$\mathbf{y} \in \mathbb{R}^{n}$  & Ground truth label vector
\\
$\hat{\mathbf{y}} \in \mathbb{R}^{n}$  & Predicted cluster-ID vector
\\
$\mathbf{X} \in \mathbb{R}^{n\times d'}$  & Node attribute matrix  
\\
$\mathbf{A} \in \mathbb{R}^{n\times n}$  & Original adjacency matrix 
\\
$\mathbf{Z} \in \mathbb{R}^{n\times d}$   & Node embeddings 
\\
$\mathbf{C} \in \mathbb{R}^{k\times d}$   & Cluster center embeddings 
\\
$\mathbf{Q} \in \mathbb{R}^{n\times k}$   & Soft clustering assignment matrix
\\
$\mathbf{P} \in \mathbb{R}^{n\times k}$    & High confidence clustering assignment matrix
\\
\bottomrule
\end{tabular}
}
\label{NOTATION_TABLE} 
\end{table}

The target of deep graph clustering is to encode nodes in the graph with the neural networks and divide them into different clusters. Figure \ref{overall} demonstrates this process. In formulaic, the self-supervised neural network $\mathcal{F}$ first encodes the nodes of the attribute graph $\mathcal{G}_{\text{A}}$ as follows.
\begin{equation} 
\mathbf{Z} = \mathcal{F}(\mathcal{G}_{\text{A}})=\mathcal{F}(\mathbf{X}, \mathbf{A}),
\label{encoding}
\end{equation}
where $\mathbf{X} \in \mathbb{R}^{n\times d'}$ denotes the node attribute matrix, $\mathbf{A}\in \mathbb{R}^{n \times n}$ denotes the adjacency matrix, and $\mathbf{Z}\in \mathbb{R}^{n \times d}$ denotes the learned node embeddings. Generally, the self-supervised neural network $\mathcal{F}$ is trained with pre-text tasks like structure reconstruction, attribute reconstruction, contrastive learning, adversarial learning, etc. Details about network architecture and learning paradigm are introduced in Section \ref{network_architecture_section} and Section \ref{learning_paradigm_section}, respectively. After encoding, the clustering method $\mathcal{C}$ groups the nodes into several disjoint clusters as follows.

\begin{figure*}[!t]
\centering
\small
\begin{minipage}{0.9\linewidth}
\centerline{\includegraphics[width=1\textwidth]{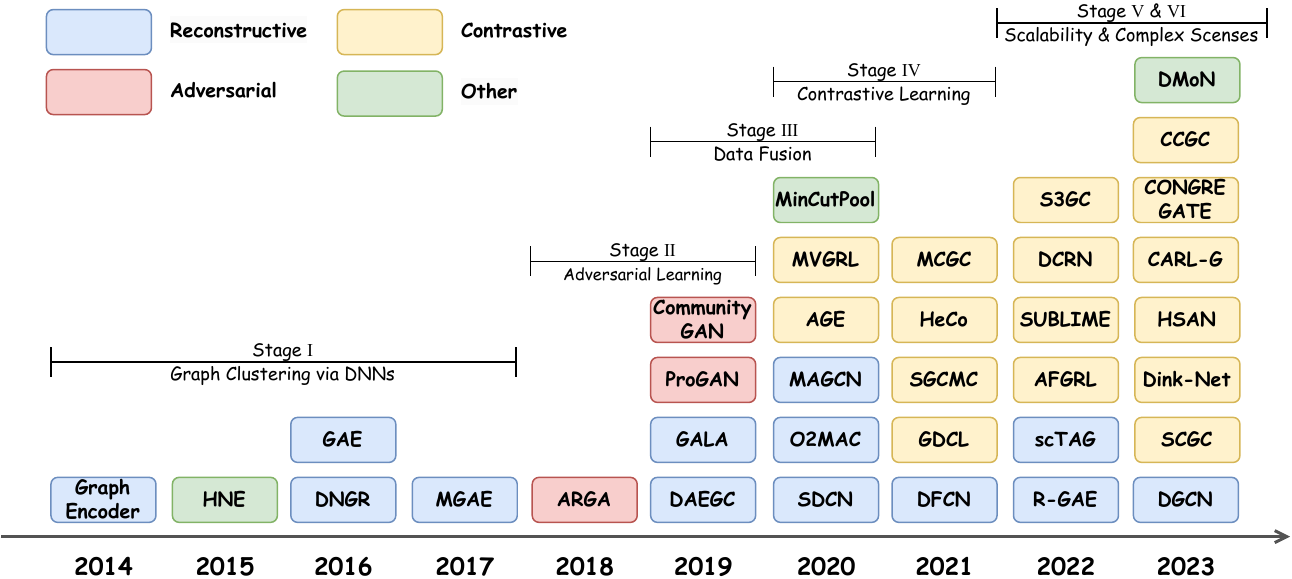}}
\end{minipage}
\caption{Time line of the important baselines in deep graph clustering field.}
\label{baseline}
\end{figure*}

\begin{equation} 
\hat{\mathbf{y}} = \mathcal{C}(\mathbf{Z}, k),
\label{clustering}
\end{equation}
where $k$ denotes the cluster number and $\hat{\mathbf{y}} \in \mathbb{R}^{n}$ denotes predicted cluster-ID vector. We deliberate the details of the clustering method $\mathcal{C}$ in Section \ref{clustering_method_section}. After the clustering process, evaluation method $\mathcal{M}$ tests the clustering performance with various metrics as follows. 
\begin{equation} 
s = \mathcal{M}(\mathbf{y}, \hat{\mathbf{y}}),
\label{evaluate}
\end{equation}
where $s \in \mathbb{R}$ denotes the clustering score and $\mathbf{y} \in \mathbb{R}^{n}$ denotes the ground truth label vector. In the next section, we introduce the development of deep graph clustering.

% The mainstream clustering methods include $k$-Means \cite{K-means}, spectral clustering \cite{spectral_clustering}, clustering neural network \cite{SDCN}, etc.

% \subsection{Dataset}
% \subsection{Dataset \& Metric}
% The benchmark graph datasets include CORA, CITESEER, PUBMED, AMAP, AMAC, etc. The detailed statistical information can be found in Appendix. To comprehensively validate the clustering performance, widely-used metrics include ACC, NMI, ARI, and F1. The detailed calculation process is formulated in Appendix. 

% Important Baselines
\subsection{Development}
\textbf{Overview}. In this section, we introduce the development of deep graph clustering methods. At first, as shown in Figure \ref{baseline}, we summarize the essential baselines in the deep graph clustering field. These papers promote the crucial development of deep graph clustering and gradually become the milestones in this field. Additionally, these research papers have been published in influential international conferences or high-quality journals in artificial intelligence, machine learning, data mining, computer vision, multimedia, etc. In Figure \ref{baseline}, the important baselines are displayed according to the publish time. Besides, in order to highlight the development trend of deep graph clustering, we roughly categorize the methods into four classes according to the learning paradigm. The blue, red, yellow, and green boxes indicate the reconstructive methods, adversarial methods, contrastive methods, and other methods. We find that most of the early methods are based on the reconstructive and adversarial learning paradigms. Recently, contrastive deep graph clustering methods become popular and mainstream. Next, we introduce the development of deep graph clustering methods in detail.

% artificial intelligence, machine learning, 

% These papers witness the 

% represent the milestone of deep graph clustering.  

% Besides, until this survey is finished, these articles are at least cited by ten times per year, and the corresponding code is available and solid. 

% Next, we introduce these important baselines in detail. 

% Figure \ref{baseline} summarizes the important baselines in deep graph clustering. 

% , e.g., AAAI, ACM MM, CIKM, ICCV, ICML, IEEE TCYB, IEEE TKDE, IEEE TMM, IEEE TNNLS, IJCAI, NeurIPS, SIGKDD, WSDM, and WWW (sorted in the alphabetical order)
% open-source 

\textbf{Stage $\text{\uppercase\expandafter{\romannumeral1}}$: graph clustering via deep neural networks}. At the early stage, motivated by the great success of deep learning, researchers aim to endow the graph clustering methods with strong representation capability of deep neural networks. Concretely, the pioneers \cite{GraphEncoder} adopt the sparse auto-encoder \cite{sparse_autoencoder} to learn the non-linear node representations and then perform $k$-means \cite{K-means} to separate the node embeddings into disjoint clusters in the GraphEncoder model. After that, DNGR \cite{DNGR} is developed to capture graph structural information by the random surfing model. Although verified effectiveness, the previous methods merely learn the graph structure information while ignoring the node attributes. GAE/VGAE \cite{GAE} is proposed with the graph convolutional encoder \cite{GCN} and a simple inner product decoder to learn both attribute and structural information. Meanwhile, to handle the heterogeneous graphs, \cite{HNE} develop a deep heterogeneous graph embedding algorithm termed HNE for five downstream tasks. Motivated by GAE/VGAE \cite{GAE}, \cite{MGAE} propose MGAE to learn node representations with the graph auto-encoder and then group the nodes into clusters with the spectral clustering algorithm \cite{spectral_clustering}.

\textbf{Stage $\text{\uppercase\expandafter{\romannumeral2}}$: introduce adversarial mechanism}. Subsequently, motivated by the generative adversarial mechanism \cite{GAN}, several adversarial deep graph clustering methods are proposed. For example, \cite{ARGA_conf,ARGA} propose ARGA/ARVGA by enforcing the latent representations to align a prior distribution in an adversarial learning manner. Similarly, ProGAN \cite{ProGAN} and CommnityGAN \cite{CommunityGAN} also utilize generative adversarial networks to generate proximities and optimize embeddings, respectively.

\textbf{Stage $\text{\uppercase\expandafter{\romannumeral3}}$: unified framework \& data fusion}. Although promising performance has been achieved, \cite{DAEGC} indicates that the previous methods are not designed for the specific clustering task. To design a clustering-directed method, \cite{DAEGC} propose a unified framework termed DAEGC with the attention-based graph encoder and clustering alignment loss adopted in deep clustering methods \cite{DEC,IDEC}. In the same year, \cite{GALA} designed a new symmetric graph auto-encoder architecture named GALA based on the Laplacian sharpening. Besides, \cite{AGC} proposes AGC model with an adaptive graph convolution to capture the clustering information in different neighbor hops. Subsequently, SDCN \cite{SDCN} and DFCN \cite{DFCN} verify the effectiveness of integrating structural information and attribute information by the delivery operator and the information fusion module, respectively. Then, to avoid the expensive costs of spectral clustering, \cite{mincutpool} formulate a continuous relaxation of the normalized minCUT problem and optimize the clustering objective with the GNN. It is worth mentioning that some graph pooling methods \cite{DMoN,diffpool,SAG} also contribute to this field. Then, O2MAC \cite{O2MAC} and MAGCN \cite{MAGCN} make attempts to exploit deep neural networks for attributed multi-view graph clustering via the multiple graph view reconstruction and the view-consistency information learning, respectively. Besides, a self-supervised method SGCMC \cite{SGCMC} utilizes the clustering labels to guide the network learning, thus improving clustering performance. R-GAE \cite{R-GAE} rethink the graph-auto-encoder-based deep graph clustering methods by considering feature randomness and feature drift. And scTAG \cite{scTAG} apply the graph-auto-encoder-based methods to the single-cell RNA sequencing. 

\textbf{Stage $\text{\uppercase\expandafter{\romannumeral4}}$: apply contrastive learning}. More recently, contrastive learning has become the mainstream paradigm in the domains of vision \cite{DIM,MOCO,SIMCLR,BYOL,BARLOW,xihong,dealmvc} and graph \cite{DGI,GRACE,simgrace,GraphCL,BYOL_graph,KGESymCL,GCL_study}, and contrastive deep graph clustering methods are increasingly proposed. Specifically, in the AGE model, \cite{AGE} first filters the high-frequency noises in node attributes and then trains the encoder by adaptively discriminating the positive and negative samples. In the same year, MVGRL \cite{MVGRL} generates an augmented structural view and contrasts node embeddings from one view with graph embeddings of another view and vice versa. Following up, MCGC \cite{MCGC} and HeCo \cite{HeCo} extend the contrastive paradigm to multi-view clustering and heterogeneous graph learning.

\textbf{Stage $\text{\uppercase\expandafter{\romannumeral5}}$: improve contrastive learning}. Although the effectiveness of the contrastive learning paradigm has been verified, there are still many open technical problems. Specifically, \cite{GDCL} proposes GDCL to correct sampling bias in contrastive deep graph clustering. Besides, to avoid the semantic drift caused by inappropriate data augmentations, AFGRL \cite{AFGRL} is proposed by replacing data augmentations with node discovery. Differently, Liu $et~al.$ \cite{SCGC} propose an augmentation-free contrastive deep graph clustering method by designing the parameter-unshared encoders. Then, to refine the noisy connections in the graph, \cite{SUBLIME} propose SUBLIME by generating the sketched graph view with unsupervised structure learning. Moreover, \cite{DCRN,IDCRN} design the dual correlation reduction strategy in the DCRN model to alleviate the representation collapse problem in deep graph clustering. To further enhance the discriminative capability of networks, Liu $et~al.$ \cite{HSAN} guide the networks to learn the hard sample pairs. CCGC \cite{CCGC} propose a new positive and negative sample pairs construction method.

\textbf{Stage $\text{\uppercase\expandafter{\romannumeral6}}$: scale to large graphs \& apply to more complex scenarios}. However, previous methods fail to scale to the large graphs, easily leading to the out-of-memory and long running time problem. To this end, Devvrit $et~al.$ \cite{S3GC} scale the contrastive deep graph clustering to large-scale graphs. Additionally, Dink-Net \cite{dink_net} unified the representation learning and clustering optimization into an end-to-end framework on large-scale graphs via dilation and shrink cluster loss functions. Besides, Shiao $et~al.$ \cite{CARL-G} utilize the node clustering method to accelerate graph representation learning. Besides, Sun $et~al.$ \cite{CONGREGATE} rethink the graph clustering problem from a geometric perspective and introduce the heterogeneous curvature space into deep graph clustering. To adapt deep graph clustering methods to both homophily and heterophily graphs, DGCN \cite{DGCN} is proposed by designing the mixed graph filter and the dual encoders. And Wen $et~al.$ \cite{wenyi} propose a framework to match the unpaired multi-view graphs.

% in complex scenarios

\subsection{Evaluation} 
This section introduces the evaluation of deep graph clustering. Deep graph clustering is a purely unsupervised task. Therefore it is hard to evaluate the clustering performance of deep graph clustering methods without the ground truth. Generally, an excellent deep graph clustering algorithm can learn the clustering distribution that has small with-cluster variance and significant between-cluster variance. The mainstream metrics can be categorized into two classes \cite{clustering_metrics}, i.e., extrinsic metrics and intrinsic metrics. Calculating extrinsic metrics requires the ground truth while calculating the intrinsic metrics does not require labels. In the research track, researchers conduct experiments on graph data with human annotations, thus the extrinsic metrics of clustering are more common. However, in the industry scenario, labels of nodes are always scarce, therefore the intrinsic metrics are more practical.

Firstly, we introduce four extrinsic metrics, including accuracy (ACC) \cite{ACC}, normalized mutual information (NMI) \cite{NMI}, adjust rand index (ARI) \cite{ARI}, and F1 score. These metrics need the ground truth labels, therefore the typical pre-process operation is to map the cluster-ID to the class-ID by the Kuhn-Munkres algorithm \cite{Kuhn-Munkres}. Then the metrics can be calculated. Specifically, accuracy (ACC) is calculated as follows:
\begin{equation}
\text{ACC}=\sum_{i=1}^n \frac{\phi({y}_{i},\text{map}(\hat{y}_i))}{n}, 
\label{ACC}
\end{equation}
where $\hat{y}_i$ and $y_i$ denote the predicted cluster-ID and the ground truth label of the $i$-th sample, respectively. Besides, $\text{map}(\cdot)$ denotes the Kuhn-Munkres algorithm \cite{Kuhn-Munkres}, which maps the predicted cluster-ID to the class-ID. $\phi(\cdot)$ is an indicator function as formulated:
\begin{equation}
\phi(y_i, \text{map}(\hat{y}_i)) = 
\left \{
\begin{aligned}
\ 1 \ \ & \text{if} \ y_i=\text{map}(\hat{y}_i), \\
\ 0 \ \ & \text{otherwise}.
\end{aligned}
\right.
\end{equation}
In addition, the F1-score also evaluates the clustering performance. It is a balance of precision and recall, calculated as follows. 
\begin{equation}
\text{F1}=2 \cdot \frac{\text{precision} \cdot \text{recall}}{\text{precision} + \text{recall}},
\label{F1}
\end{equation}

\begin{equation}
\text{precision} = \frac{\text{TP}}{\text{TP}+\text{FP}}, \ \text{recall} = \frac{\text{TP}}{\text{TP}+\text{FN}},
\label{precision}
\end{equation}
where TP, FP, and FN denote true positive samples, false positive samples, and false negative samples, respectively. Further, the calculation of Normalized Mutual Information (NMI) is based on mutual information. It is more robust to the unbalanced sample distribution. It can be formulated as follows. 
\begin{equation}
% \small
\text{NMI}=-\frac{2\sum_{\hat{y}}\sum_y p(\hat{y},y) \log \frac{p(\hat{y},y)}{p(\hat{y})p(y)}}{\sum_i p(\hat{y}_i) \log(p(\hat{y}_i))+\sum_j p(y_j) \log(p(y_j))},
\label{NMI}
\end{equation}
where $p(\hat{y})$, $p(y)$, and $p(\hat{y},y)$ denotes the distribution of the predicted results, distribution of the ground truth, and joint distribution of them, respectively. Differently, another metric named Adjust Rand Index (ARI) is based on the pairwise similarity between predicted labels and ground truth labels. It is formulated as follows: 
\begin{equation}
\text{ARI} = \frac{\text{RI}-\text{expectedRI}}{\max(\text{RI}) - \text{expectedRI}},
\end{equation}
where RI denote rand index \cite{ARI} and expectedRI \cite{ARI} denote the expected rand index. ARI equals 0 indicates real and modeled clustering do not agree on pairing, and ARI equals 1 indicates real and modeled clustering both represent the same clusters.
 
% \begin{equation}
% \small
% \text{ARI} = \frac{\sum_{i j}\tbinom{n_{i j}}{2} - \left[\sum_{i} \tbinom{a_{i}}{2}\sum_{j} \tbinom{b_{j}}{2}\right] / \tbinom{n}{2}}{\frac{1}{2}\left[\sum_{i} \tbinom{a_{i}}{2}+\sum_{j} \tbinom{b_{j}}{2} \right]-{\left[ \sum_{i} \tbinom{a_{i}}{2} \sum_{j} \tbinom{b_{j}}{2} \right] / \tbinom{n}{2}}},
% \end{equation}

% ARI is based on the similarity of pairwise labels between the ground truth and predicted results as formulated:

Secondly, we introduce three intrinsic metrics, including Silhouette Coefficient (SC) \cite{SC}, Calinski-Harabasz Index (CHI) \cite{CHI}, and Davies-Bouldin Index (DBI) \cite{DBI}. Different from previous extrinsic metrics, the intrinsic metrics are more piratical for real-world data since they get rid of the ground truth labels. Concretely, Silhouette Coefficient (SC) can be calculated as follows. 
\begin{equation}
\text{SC} = \frac{dis-dis'}{\max(dis, dis')},
\label{SC}
\end{equation}
where $dis'$ denotes the average distance between the node and all other nodes in the same cluster while $dis$ denotes the average distance between the node and all other points in the next nearest cluster. SC is intuitive and easy to understand and has a range of $[-1, 1]$. Additionally, Calinski-Harabasz Index can be calculated as follows. 
\begin{equation}
b = \sum_{i=1}^{k} \mathbf{m}_i (\mathbf{c}_i-\hat{\mathbf{c}})(\mathbf{c}_i- \hat{\mathbf{c}})^\text{T},
\label{B}
\end{equation}

\begin{equation}
w =  \sum_{i=1}^{k}  \sum_{\mathbf{x} \in i\text{-th} \ \text{cluster}} (\mathbf{x}-\mathbf{c}_i)(\mathbf{x}-\mathbf{c}_{i})^\text{T},
\label{W}
\end{equation}

\begin{equation}
\text{CHI} = \frac{b}{w} \times \frac{n-k}{k-1},
\label{CHI}
\end{equation}
where $b$ denotes the between-cluster dispersion, and $w$ denotes the within-cluster dispersion. $n$ is the number of all nodes, $k$ is the cluster number, and $\mathbf{m}_i$ is the number of nodes in the $i$-th cluster. Besides, $\hat{\mathbf{c}}$ denote the cluster center of all samples, and $\mathbf{c}_i$ denote the $i$-th cluster center. CHI measures the between-cluster dispersion against within-cluster ones. The higher score indicates better clusters. Unlike CHI, Davies-Bouldin Index (DBI) can measure the cluster size against the average distance between clusters. And the lower score indicates better clusters. DBI can be calculated as follows. 
\begin{equation}
r_{ij} = \frac{s_i+s_j}{s'_{ij}}, \  \text{DBI} = \frac{1}{k} \sum_{i=1}^{k} \max_{i\neq j} r_{ij},
\label{Rij}
\end{equation}
where $s_i$ denotes the average distance between each nodes in the $i$-th cluster to the $i$-th cluster center $\mathbf{c}_i$. $s'_{ij}$ denotes the distance between the $i$-th cluster center $\mathbf{c}_i$ to the $j$-th cluster center $\mathbf{c}_j$. In this paper, the used datasets are open benchmarks, which have the ground truth of the nodes. Therefore, we adopt four extrinsic metrics, including ACC, NMI, ARI, and F1 score in our experiments.

\section{Taxonomy}
\label{taxonomy}

We contribute a structured taxonomy to provide a broad overview of this field. Concretely, this section introduces the taxonomy of deep graph clustering methods from the following four perspectives: graph type, network architecture, learning paradigm, and clustering method. The surveyed paper are categorized based these criteria in Table \ref{taxonomy_table_1} (part $ \text{\uppercase\expandafter{\romannumeral1}}$) and Table \ref{taxonomy_table_2} (part $ \text{\uppercase\expandafter{\romannumeral2}}$). Next, we present taxonomy criteria in detail.

\begin{table*}[!t]
\centering
\caption{Taxonomy table of deep graph clustering methods: part $ \text{\uppercase\expandafter{\romannumeral1}}$. The criteria contain four aspects: graph type, network architecture, learning paradigm, and clustering method.}
\resizebox{0.95\linewidth}{!}{
% \scalebox{1.0}{
\begin{tabular}{ccccc} 
\hline
\textbf{Methods} & \textbf{Graph Type}  & \textbf{Network Architecture} & \textbf{Learning Paradigm} & \textbf{Clustering Method}  \\ 
\hline
GraphEncoder \cite{GraphEncoder}     & Pure Structure Graph & MLP                           & Reconstructive             & Traditional Clustering      \\
DNGR   \cite{DNGR}          & Pure Structure Graph & MLP                           & Reconstructive             & Traditional Clustering      \\
CommunityGAN  \cite{CommunityGAN}   & Pure Structure Graph & MLP                           & Adversarial                & Neural Clustering           \\
NetVAE   \cite{NetVAE}        & Attribute Graph      & MLP                           & Reconstructive             & Traditional Clustering      \\
DGCN   \cite{DGCN}        & Attribute Graph      & MLP                           & Reconstructive             & Neural Clustering      \\
GRACE   \cite{GRACE_yang}        & Attribute Graph      & MLP                           & Reconstructive             & Neural Clustering      \\
ProGAN    \cite{ProGAN}       & Attribute Graph      & MLP                           & Adversarial                & Traditional Clustering      \\
AGE      \cite{AGE}        & Attribute Graph      & MLP                           & Contrastive                & Traditional Clustering      \\
SCGC      \cite{SCGC}        & Attribute Graph      & MLP                           & Contrastive                & Traditional Clustering      \\
CCGC      \cite{CCGC}        & Attribute Graph      & MLP                           & Contrastive                & Traditional Clustering      \\
HSAN      \cite{HSAN}        & Attribute Graph      & MLP                           & Contrastive                & Traditional Clustering      \\
RGC      \cite{liuyue_RGC}        & Attribute Graph      & MLP                           & Contrastive                & Traditional Clustering      \\
CONVERT      \cite{convert}        & Attribute Graph      & MLP                           & Contrastive                & Traditional Clustering      \\
GCC-LDA      \cite{GCC-LDA}        & Attribute Graph      & MLP                           & Contrastive                & Traditional Clustering      \\
GALA     \cite{GALA}        & Attribute Graph      & GNN                           & Reconstructive             & Traditional Clustering      \\
MGAE     \cite{MGAE}        & Attribute Graph      & GNN                           & Reconstructive             & Traditional Clustering      \\
CGCN     \cite{CGCN}        & Attribute Graph      & GNN                           & Reconstructive             & Traditional Clustering      \\
GCLN     \cite{GCLN}        & Attribute Graph      & GNN                           & Reconstructive             & Traditional Clustering      \\
AHGAE    \cite{AHGAE}        & Attribute Graph      & GNN                           & Reconstructive             & Traditional Clustering      \\
RWR-GAE    \cite{RWR-GAE}        & Attribute Graph      & GNN                           & Reconstructive             & Traditional Clustering      \\
EGAE     \cite{EGAE}        & Attribute Graph      & GNN                           & Reconstructive             & Traditional Clustering      \\
DGCSF     \cite{DGCSF}        & Attribute Graph      & GNN                           & Reconstructive             & Traditional Clustering      \\
DAEGC    \cite{DAEGC}        & Attribute Graph      & GNN                           & Reconstructive             & Neural Clustering           \\
DGVAE    \cite{DGVAE}        & Attribute Graph      & GNN                           & Reconstructive             & Neural Clustering           \\
CDRS      \cite{CDRS}       & Attribute Graph      & GNN                           & Reconstructive             & Neural Clustering           \\
GCC       \cite{GCC}       & Attribute Graph      & GNN                           & Reconstructive             & Neural Clustering           \\
GC-SEE       \cite{GC-SEE}       & Attribute Graph      & GNN                           & Reconstructive             & Neural Clustering           \\
FT-VGAE      \cite{FT-VGAE}       & Attribute Graph      & GNN                           & Reconstructive             & Neural Clustering           \\
scTAG      \cite{scTAG}       & Attribute Graph      & GNN                           & Reconstructive             & Neural Clustering           \\
GC-VAE      \cite{GC-VAE}       & Attribute Graph      & GNN                           & Reconstructive             & Neural Clustering           \\
DNENC      \cite{DNENC}       & Attribute Graph      & GNN                           & Reconstructive             & Neural Clustering           \\
JANE      \cite{JANE}       & Attribute Graph      & GNN                           & Adversarial                & Traditional Clustering      \\
SAIL       \cite{SAIL}      & Attribute Graph      & GNN                           & Contrastive                & Traditional Clustering      \\
AFGRL      \cite{AFGRL}      & Attribute Graph      & GNN                           & Contrastive                & Traditional Clustering      \\
S$^3$GC      \cite{S3GC}       & Attribute Graph      & GNN                           & Contrastive                & Traditional Clustering      \\
CONGREGATE      \cite{CONGREGATE}       & Attribute Graph      & GNN                           & Contrastive                & Traditional Clustering      \\
SCAGC   \cite{SCAGC}         & Attribute Graph      & GNN                           & Contrastive                & Neural Clustering           \\
CommDGI    \cite{CommDGI}      & Attribute Graph      & GNN                           & Contrastive                & Neural Clustering           \\
CARL-G    \cite{CARL-G}      & Attribute Graph      & GNN                           & Contrastive                & Neural Clustering           \\
\hline
\end{tabular}}
\label{taxonomy_table_1}
\end{table*}

\begin{table*}[!t]
\centering
\caption{Taxonomy table of deep graph clustering methods: part $ \text{\uppercase\expandafter{\romannumeral2}}$. The criteria contain four aspects: graph type, network architecture, learning paradigm, and clustering method.}
\resizebox{0.95\linewidth}{!}{
% \scalebox{1.0}{
\begin{tabular}{ccccc} 
\hline
\textbf{Methods} & \textbf{Graph Type}  & \textbf{Network Architecture} & \textbf{Learning Paradigm} & \textbf{Clustering Method}  \\ 
\hline
AGAE      \cite{AGAE}       & Attribute Graph      & GNN                           & Reconstructive+Adversarial & Traditional Clustering      \\
ARGA    \cite{ARGA}         & Attribute Graph      & GNN                           & Reconstructive+Adversarial & Traditional Clustering      \\
WARGA    \cite{WARGA}         & Attribute Graph      & GNN                           & Reconstructive+Adversarial & Traditional Clustering      \\
GCI    \cite{GCI}         & Attribute Graph      & GNN                           & Reconstructive+Adversarial & Traditional Clustering      \\
NCAGC    \cite{NCAGC}         & Attribute Graph      & GNN                           & Reconstructive+Contrastive & Neural Clustering      \\
GEC-CSD    \cite{GEC-CSD}         & Attribute Graph      & GNN                           & Reconstructive+Adversarial & Neural Clustering      \\
SDCN     \cite{SDCN}        & Attribute Graph      & MLP+GNN                       & Reconstructive             & Neural Clustering           \\
DFCN      \cite{DFCN}       & Attribute Graph      & MLP+GNN                       & Reconstructive             & Neural Clustering           \\
AGCN     \cite{AGCN}        & Attribute Graph      & MLP+GNN                       & Reconstructive             & Neural Clustering           \\
R-GAE     \cite{R-GAE}        & Attribute Graph      & MLP+GNN                       & Reconstructive             & Neural Clustering           \\
DAGC     \cite{DAGC}        & Attribute Graph      & MLP+GNN                       & Reconstructive             & Neural Clustering           \\
MVGRL     \cite{MVGRL}       & Attribute Graph      & MLP+GNN                       & Contrastive                & Traditional Clustering      \\
SUBLIME    \cite{SUBLIME}      & Attribute Graph      & MLP+GNN                       & Contrastive                & Traditional Clustering      \\
GDCL      \cite{GDCL}       & Attribute Graph      & MLP+GNN                       & Contrastive                & Neural Clustering           \\
Dink-Net      \cite{dink_net}       & Attribute Graph      & MLP+GNN                       & Contrastive                & Neural Clustering           \\
SELENE      \cite{SELENE}       & Attribute Graph      & MLP+GNN                       & Reconstructive+Contrastive & Traditional Clustering           \\
DCRN      \cite{DCRN}       & Attribute Graph      & MLP+GNN                       & Reconstructive+Contrastive & Neural Clustering           \\
IDCRN      \cite{IDCRN}       & Attribute Graph      & MLP+GNN                       & Reconstructive+Contrastive & Neural Clustering           \\
SCGC      \cite{SCGC_kula}       & Attribute Graph      & MLP+GNN                       & Reconstructive+Contrastive & Neural Clustering           \\
AGC-DRR     \cite{AGC-DRR}     & Attribute Graph      & MLP+GNN                       & Adversarial+Contrastive    & Neural Clustering           \\
DMGC      \cite{DMGC}       & Heterogeneous Graph  & MLP                           & Reconstructive             & Traditional Clustering      \\
MAGCN     \cite{MAGCN}       & Heterogeneous Graph  & GNN                           & Reconstructive             & Neural Clustering           \\
O2MAC     \cite{O2MAC}       & Heterogeneous Graph  & GNN                           & Reconstructive             & Neural Clustering           \\
HeCo       \cite{HeCo}      & Heterogeneous Graph  & GNN                           & Contrastive                & Traditional Clustering      \\
SGCMC     \cite{SGCMC}       & Heterogeneous Graph  & GNN                           & Reconstructive+Contrastive & Neural Clustering           \\
VaCA-HINE  \cite{vaca-hine}      & Heterogeneous Graph  & GNN                           & Reconstructive+Contrastive & Traditional Clustering      \\
HNE      \cite{HNE}        & Heterogeneous Graph  & MLP+CNN                       & Reconstructive             & Traditional Clustering      \\
VGRGMM       \cite{VGRGMM}    & Dynamic Graph        & GNN                           & Reconstructive             & Traditional Clustering      \\
TGC       \cite{TGC}    & Dynamic Graph        & MLP                           & Reconstructive             & Neural Clustering       \\
CGC       \cite{CGC}       & Dynamic Graph        & GNN                           & Contrastive                & Traditional Clustering      \\
\hline
\end{tabular}}
\label{taxonomy_table_2}
\end{table*}
% TGC

% As shown in Figure \ref{graph_type}, the input graphs of the existing deep graph clustering methods are categorized into four types: pure structure graph, attribute graph, heterogeneous graph, and dynamic graph. 

\subsection{Graph Type} \label{graph_type_section}
\begin{figure}[h]
\centering
\small
\begin{minipage}{0.9\linewidth}
\centerline{\includegraphics[width=1\textwidth]{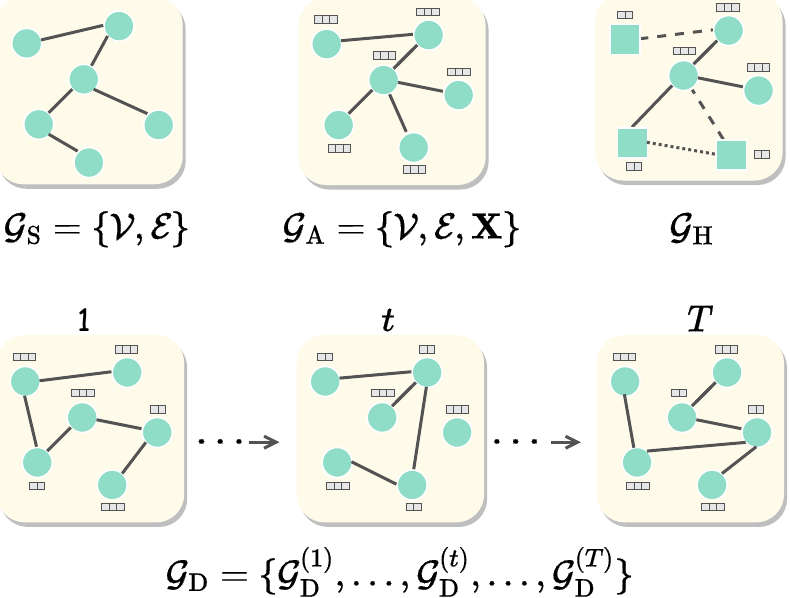}}
\end{minipage}
\caption{The demonstrations of pure structure graph $\mathcal{G}_\text{S}$, attribute graph $\mathcal{G}_\text{A}$, heterogeneous graph $\mathcal{G}_\text{H}$, and dynamic graph $\mathcal{G}_{\text{D}}$. The pure structure graph merely contains the connected edges between nodes. In addition, each node in the attribute graph attaches attributes. Moreover, in heterogeneous graphs, the node types and edge types are multiple. Furthermore, the nodes and edges of dynamic graphs will change over time.}
\label{graph_type}
\end{figure}

First of all, we begin with the input of deep graph clustering. The input graphs of existing deep graph clustering methods are mainly categorized into four types. Figure \ref{graph_type} demonstrates the details of these four types of graphs. In what follows, we provide the detailed definitions of these graph types.

\subsubsection{Pure Structure Graph} 
In a pure structure graph $\mathcal{G}_{\text{S}}=\{\mathcal{V}, \mathcal{E}\}$, $\mathcal{V}=\{v_1, v_2, \dots, v_n\}$ denotes the node set, which contain $n$ nodes of $k$ categories. Besides, $\mathcal{E}$ denotes the edge set, which contains $l$ edges. With the matrix form, the pure structure graph can be represented by the adjacency matrix $\mathbf{A}\in \mathbb{R}^{n \times n}$. Here, $a_{ij}=1$ indicates that $i$-th node is linked to $j$-th node while $a_{ij}=0$ indicates that $i$-th node is not linked to $j$-th node in the graph.

\subsubsection{Attribute Graph} 
Compared with the pure structure graph $\mathcal{G}_{\text{S}}$, the attribute graph $\mathcal{G}_{\text{A}} = \{\mathcal{V}, \mathcal{E}, \mathbf{X}\}$ is more complex. Specifically, each node in $\mathcal{G}_{\text{A}}$ attaches the $d'$-dimension attributes. $\mathbf{X} \in \mathbb{R}^{n \times d'}$ is denoted as the node attribute matrix, where $d'$ is the dimension number of node attributes. In the matrix form, $\mathcal{G}_{\text{A}}$ is represented by the adjacency matrix $\mathbf{A} \in \mathbb{R}^{n \times n}$ and node attribute matrix $\mathbf{X}$.

% \begin{definition}
% \mathbf{Attribute Graph}. Compared with the pure structure graph $\mathcal{G}_{\text{S}}$, the attribute graph $\mathcal{G}_{\text{A}} = \{\mathcal{V}, \mathcal{E}, \mathbf{X}\}$ attaches the attributes of nodes. $\mathbf{X} \in \mathbb{R}^{N \times D}$ is denoted as the attribute matrix, where $D$ is the dimension number of node attributes. With matrix form, $\mathcal{G}_{\text{A}}$ is represented by the adjacency matrix $\mathbf{A} \in \mathbb{R}^{N \times N}$ and attribute matrix $\mathbf{X}$.
% \end{definition}

\subsubsection{Heterogeneous Graph}
% The previews two type of graphs contain one type nodes and one type edges. 

The heterogeneous graph at least contains more than one type of node or more than one type of edge. Formally, we first denote the type number of nodes and edges as $T_\text{n}$ and $T_\text{e}$. Then a heterogeneous graph $\mathcal{G}_{H}$ satisfies $T_{\text{n}}+T_{\text{e}}>2$, i.e., it contains multiple types of nodes or/and multiple types of edges. Otherwise, the graph is homogeneous.

% \begin{definition}
% \mathbf{Heterogeneous Graph}. In a graph, we firstly denote the type number of nodes and edges as $T_{\text{n}}$ and $T_{\text{e}}$. A heterogeneous graph $\mathcal{G}_{\text{H}}$ satisfies $T_{\text{n}}+T_{\text{e}}>2$, i.e., it contains multiple types of nodes or/and multiple types of edges. Otherwise, the graph is a homogeneous graph.
% \end{definition}

\subsubsection{Dynamic Graph}
Different from the static graph, the dynamic graph $\mathcal{G}_{\text{D}}=\{\mathcal{G}_{\text{D}}^{(1)}, ..., \mathcal{G}_{\text{D}}^{(t)}, ..., \mathcal{G}_{\text{D}}^{(T)}\}$ will dynamically change over time, where $t$ denotes the time step. The changes in dynamic graphs include changing the linkages between nodes, adding or deleting nodes, modifying the node features, etc.

% nodes itself, or node features. 

% \begin{definition}
% \mathbf{Dynamic Graph}. Different from the static graph, the dynamic graph $\mathcal{G}_{\text{D}}=\{\mathcal{G}_{\text{D}}^{(1)}, ..., \mathcal{G}_{\text{D}}^{(t)}, ..., \mathcal{G}_{\text{D}}^{(T)}\}$ will dynamically change over time, where $t$ is the time step. 
% \end{definition}

\subsubsection{Processing Techniques}
The first type of graph, the pure structure graph, is relatively easy to process since it merely contains structural information. For example, in early works, \cite{GraphEncoder} encodes the structural embeddings with the sparse auto-encoders \cite{sparse_autoencoder}. Besides, \cite{DNGR} apply random surfing in the graph and embed the graph structure with the stacked denoising auto-encoder. For attribute graphs, the additional attribute information always brings more process operations and performance improvement. For instance, \cite{MGAE} integrates the attributes and the graph structure with the graph convolutional operation \cite{GCN}. Also, \cite{SDCN} transfers the attribute embeddings to the GCN layer with the delivery operator. Moreover, \cite{DFCN,AGCN} demonstrate the effectiveness of the attribute-structure fusion mechanism. The heterogeneous graph is more complex since it may contain different types of nodes and edges. To handle these graphs, \cite{HeCo} adopts the meta-path technology while \cite{O2MAC} treats them as multi-view graphs. Different from the aforementioned static graphs, dynamic graphs change over time, increasing the difficulty of clustering. To solve this problem, \cite{CGC} performs graph clustering in an incremental learning fashion. \cite{VGRGMM} adopt the variant of gated recurrent unit (GRU) \cite{GRU} to capture the temporal information. Liu $et~al.$ \cite{TGC} present a general framework for temporal deep graph clustering and \cite{TGC_dataset} collect several datasets for temporal deep graph clustering.

\subsection{Network Architecture}\label{network_architecture_section}
For the network architecture, the mainstream deep graph clustering methods can roughly be categorized into three classes: MLP-based method, GNN-based method, and hybrid methods.

\subsubsection{MLP-based Methods} 
The MLP-based methods utilize the multi-layer perceptrons (MLPs) \cite{MLP} to exact the informative features in the graphs. For example, GraphEncoder \cite{GraphEncoder} and DNGR \cite{DNGR} encode the graph structure with the auto-encoders. Subsequently, in ProGAN \cite{ProGAN} and CommunityGAN \cite{CommunityGAN}, the authors adopt MLPs to design the generative adversarial networks. Moreover, based on MLPs, AGE \cite{AGE} and SCGC \cite{SCGC} design the adaptive encoder and the parameter un-shared encoders to embed the smoothed node features into the latent space. Although the effectiveness of these methods has been demonstrated, it is difficult for MLPs to capture non-euclidean structural information in graphs. Thus, GNN-based methods have been increasingly proposed in recent years.

% the promising performance has been achieved

\subsubsection{GNN-based Methods} 
The GNN-based methods model the non-euclidean graph data with the GNN encoders like graph convolutional network \cite{GCN}, graph attention network \cite{GAT}, graph auto-encoder \cite{GAE}, etc. Benefiting from the strong graph structure representation capability, GNN-based methods have achieved promising performance. For instance, MGAE \cite{MGAE} is proposed to learn the node attribute and graph structure with the designed graph auto-encoder. In addition, \cite{GALA} design a novel symmetric graph auto-encoder named GALA. Furthermore, the GNNs are also applied to the heterogeneous graphs in O2MAC \cite{O2MAC}, MAGCN \cite{MAGCN}, SGCMC \cite{SGCMC}, and HeCo \cite{HeCo}. However, the entanglement of transformation and propagation in GNNs will incur heavy computational overhead. Thus, SCGC \cite{SCGC} is proposed to improve the efficiency and scalability of the existing deep graph clustering methods by decoupling these two operations.

\subsubsection{Hybrid Methods} 
More recently, some researchers have considered integrating the advantages of both MLP-based and GNN-based methods. Concretely, \cite{SDCN} transfers the embeddings from the auto-encoder to the GCN layer with a designed delivery operator. In addition, AGCN \cite{AGCN} and DFCN \cite{DFCN} demonstrate the effectiveness of the fusion of the node attribute feature and the topological graph feature with the combination of auto-encoder and GCN. Moreover, some contrastive deep graph clustering methods \cite{MVGRL,GDCL,SUBLIME,DCRN,AGC-DRR,CCGC,convert,GCC-LDA} also adopt the hybrid architecture of MLPs and GNNs as the backbones.

\subsection{Learning paradigm} \label{learning_paradigm_section}
Based on the learning paradigm, the surveyed methods contain four categories: reconstructive methods, adversarial methods, contrastive methods, and hybrid methods. Taking the attribute graph as the input graph, we elaborate on these learning paradigms of deep graph clustering methods as follows.

\begin{figure}[h]
\centering
\small
\begin{minipage}{0.9\linewidth}
\centerline{\includegraphics[width=1\textwidth]{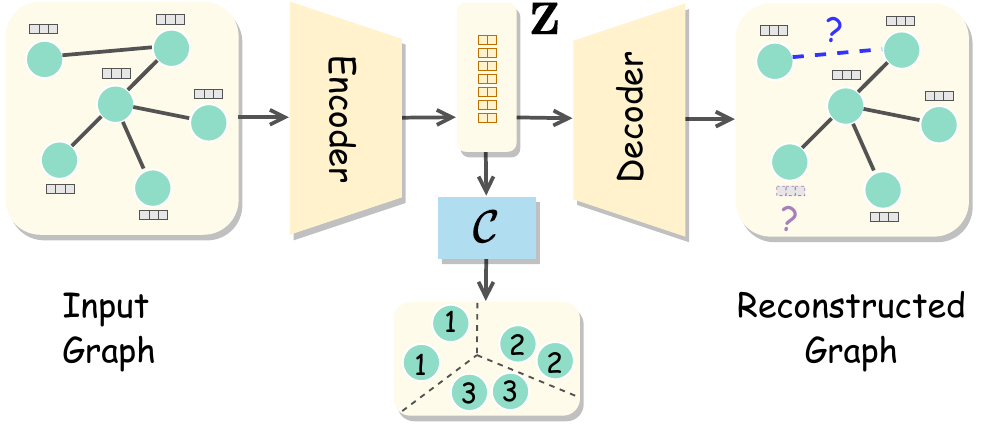}}
\end{minipage}
\caption{The general pipeline of the reconstructive deep graph clustering methods. Firstly, the nodes of the input graph are encoded into the node embeddings $\mathbf{Z}$ via a designed Encoder. Subsequently, with the reconstruction pre-text tasks like attribute reconstruction or link reconstruction, the decoder aims to recover the graph information from the learned embeddings $\mathbf{Z}$. Eventually, in the latent space, the clustering method $\mathcal{C}$ groups the nodes into different clusters.}
\label{reconstructive}
\end{figure}

\subsubsection{Reconstructive Methods} 

The core idea of the reconstructive methods is forcing the network to encode the features in the graph and then recovering (part of) the graph information from the learned embeddings. Thus, the reconstructive methods focus on the intra-data information in the graph and adopt the node attributes and graph structure as the self-supervision signals. The general pipeline of reconstructive deep graph clustering methods is illustrated in Figure \ref{reconstructive}. The core designs of the reconstructive methods are the encoder architecture, decoder architecture, and reconstructed objective functions. Researchers improve reconstructive methods from these perspectives.

% Concretely, the encoder aims to embed the nodes into the latent space, and the decoder aims to reconstruct the original graph information from the latent code. Typically, the decoder will reconstruct the node attributes or/and the graph structure. In this manner, the learned latent embeddings better represent the graph information from the original graph. 

\subsubsection{Adversarial Methods} 

\begin{figure}[h]
\centering
\small
\begin{minipage}{0.9\linewidth}
\centerline{\includegraphics[width=1\textwidth]{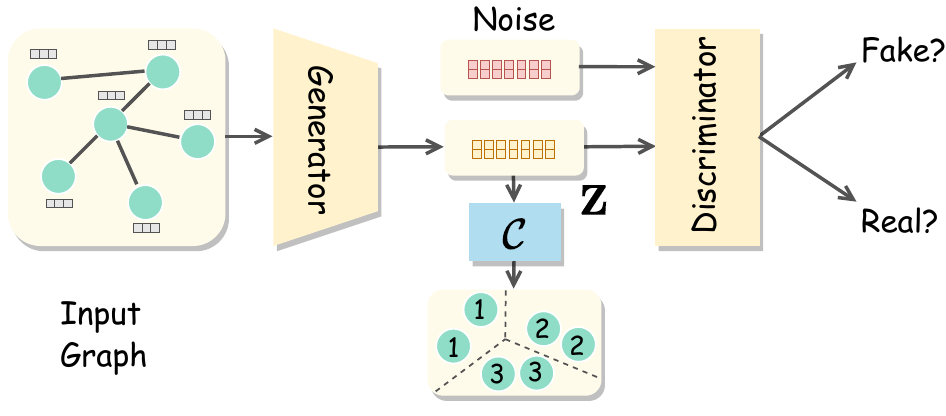}}
\end{minipage}
\caption{The general pipeline of adversarial deep graph clustering. Firstly, the Generator aims to generate high-quality node embeddings $\mathbf{Z}$ from the input graph. Subsequently, the Discriminator is trained to distinguish the fake information and the learned features. After the self-supervised training, the clustering method $\mathcal{C}$ separates the learned node embeddings into several clusters.}
\label{adversarial}
\end{figure}

The adversarial deep graph clustering methods aim to improve the quality of features by creating a zero-sum game between the generator and the discriminator. Specifically, the discriminator is trained to recognize whether learned features are from the real data distribution, while the generator aims to generate confusing embeddings to cheat the discriminator. In these settings, the generation and discrimination operations provide effective self-supervision signals, guiding the network to learn high-quality embeddings. Figure \ref{adversarial} demonstrates the general pipeline of the adversarial deep graph clustering methods. The core technologies determining the quality of adversarial methods contain generator designs, discriminator designs, noise generation, and discriminative loss functions. Several works aim to improve the performance of adversarial methods from these aspects.

% fake information and $\mathbf{Z}$ are

\subsubsection{Contrastive Methods} 

\begin{figure}[h]
\centering
\small
\begin{minipage}{0.9\linewidth}
\centerline{\includegraphics[width=1\textwidth]{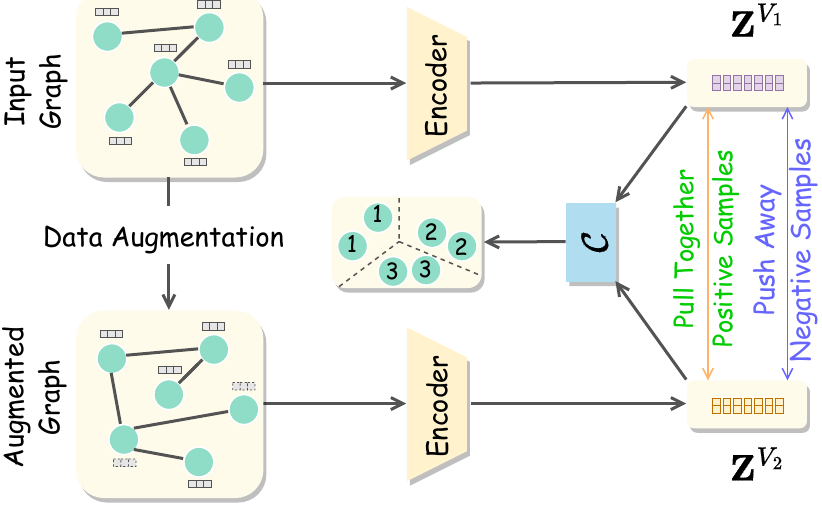}}
\end{minipage}
\caption{The general pipeline of contrastive deep graph clustering methods. At first, the augmented graph is generated via data augmentations, and nodes are embedded into the latent space by the encoders. Subsequently, the networks are guided to pull together positive samples and push away negative samples. Finally, the node embeddings from different views are fused and grouped into distinct clusters via the clustering method $\mathcal{C}$.}
\label{contrastive}
\end{figure}

The critical idea in contrastive deep graph clustering methods is to improve the discriminativeness of features by pulling together the positive samples while pushing away the negative ones. Thus, contrastive methods focus on the inter-data information and construct the self-supervision signals via meaningful relationships between samples. The general pipeline of contrastive methods can be found in Figure \ref{contrastive}. The main techniques in contrastive methods include graph data augmentations, siamese encoder designs, positive and negative sample pair construction, negative sampling, counteractive learning loss functions, etc. These aspects are carefully modified to enhance the discriminative capability of contrastive learning methods.

\subsubsection{Hybrid Methods} 

In recent years, some papers have demonstrated the effectiveness of combining different learning paradigms. For example, in ARGA \cite{ARGA}, Pan $et~al.$ adopt both the reconstructive and adversarial learning paradigms. Besides, researchers \cite{SGCMC,DCRN} also verified the effectiveness of the combination of reconstructive and contrastive learning paradigms. Moreover, in AGC-DRR \cite{AGC-DRR}, the researchers show that an adversarial mechanism is a new option for data augmentation in contrastive learning. How to better optimize multiple self-supervised tasks and make them cooperate with each other is another crucial research topic \cite{AutoSSL,multi_task}.

\subsection{Clustering Method} \label{clustering_method_section}
The clustering methods in deep graph clustering aim to separate the learned node embeddings into different clusters in a purely unsupervised manner. They roughly can be categorized into two classes: traditional clustering and neural clustering.

\subsubsection{Traditional Clustering} 
The traditional clustering methods \cite{K-means,spectral_clustering,GMM,LSWMKC-2022,traditional_1,LiangLiTKDE} can be directly performed on the learned node embeddings to group them into disjoint clusters in many early deep graph clustering methods \cite{GraphEncoder,DNGR,MGAE,ARGA,GALA,AGAE,MVGRL,AGE}. Although they efficiently achieve promising performance, the traditional clustering methods have two drawbacks as follows: 1) First, the clustering process can not benefit from the strong representation capability of deep neural networks, thus limiting the discriminative capability of the samples. 2) Second, in these methods, the node representation learning and the clustering learning cannot be jointly optimized in an end-to-end manner, therefore leading to sub-optimal performance. 3) These methods are not easy to adopt batch training and batch inference techniques, limiting the model scalability on large graphs.

\begin{figure}[h]
\centering
\small
\begin{minipage}{0.90\linewidth}
\centerline{\includegraphics[width=1\textwidth]{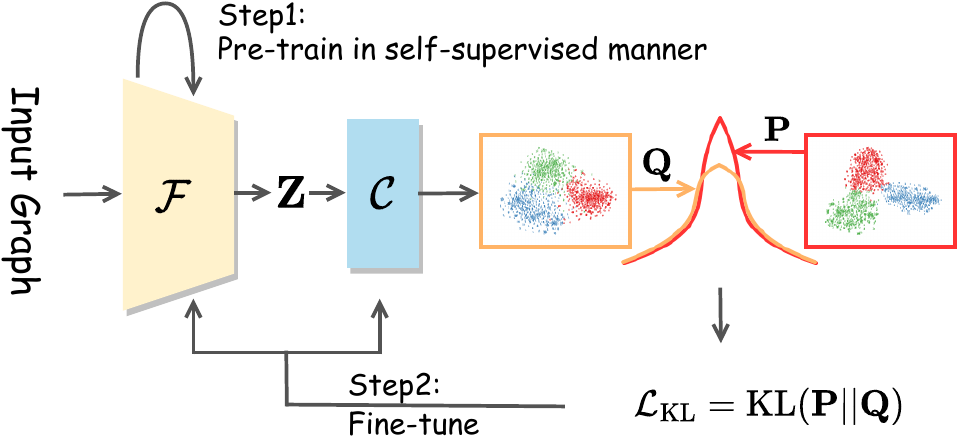}}
\end{minipage}
\caption{The clustering distribution alignment loss $\mathcal{L}_{\text{KL}}$ in the neural clustering methods. In the first step, the graph encoder $\mathcal{F}$ is trained in a self-supervised manner. Then, the clustering method obtains the initial clustering distribution $\mathbf{Q}$. In the second step, the model sharpens the clustering distribution $\mathbf{Q}$ and gains the sharpened distribution $\mathbf{P}$, which represents a clustering distribution with more confidence. The model will be fine-tuned by aligning the original distribution and sharpened distribution via the KL divergence loss function. }
\label{kl_loss_figure}
\end{figure}

\subsubsection{Neural Clustering} 

To alleviate the aforementioned issues of traditional clustering methods, the neural clustering methods are designed to group the samples into different clusters with deep neural networks. Concretely, in the neural clustering methods, the clustering process and the deep neural network are jointly optimized by the gradient descent algorithms \cite{ADAM,SGD}.

For example, in many two-stage neural clustering methods \cite{DAEGC,MAGCN,O2MAC,SDCN,AGCN,GDCL,DFCN,DCRN,CDRS}, a widely-used clustering distribution alignment loss is introduced to network optimization as shown in Figure \ref{kl_loss_figure}. Specifically, in the first stage, these methods pre-train the encoders $\mathcal{F}$ in a self-supervised manner. After that, they initialize the cluster center embeddings $\mathbf{C} \in \mathbb{R}^{k \times d}$ by the traditional clustering algorithms on the learned node embeddings $\mathbf{Z} \in \mathbb{R}^{n \times d}$, where $n$, $k$, and $d$ denotes the node number, cluster number, latent feature dimension, respectively. Note that the cluster center embeddings $\mathbf{C}$ are set to the learnable parameters of the deep neural networks, i.e., the tensors with gradients. In the second stage, a soft assignment between the node embeddings $\mathbf{Z}$ and the cluster center embeddings $\mathbf{C}$ is calculated as formulated:

\begin{equation}
Q_{ij} = \frac{(1+\|\mathbf{z}_i-\mathbf{c}_j\|^2/\alpha)^{-\frac{\alpha+1}{2}}}{\sum_{j'}(1+\|\mathbf{z}_i-\mathbf{c}_j\|^2/\alpha)^{-\frac{\alpha+1}{2}}}.
\label{soft_assignment}
\end{equation}
Here, the similarity between $i$-th node embedding $\mathbf{z}_i$ and $j$-th cluster center embedding $\mathbf{c}_j$ is measured by the Student's $t$-distribution kernel. $\alpha$ is the freedom degree of Student's $t$-distribution. $Q_{ij}$ can be considered as the probability of assigning $i$-th node to $j$-th cluster, namely a soft assignment. Subsequently, the clustering distribution is promoted by aligning the soft assignment $Q_{ij}$ with the high confidence assignments $P_{ij}$ as follows. 

\begin{equation}
\mathcal{L}_{\text{KL}} = \text{KL}(P||Q) = \sum_i \sum_j P_{ij}\log(\frac{P_{ij}}{Q_{ij}}),
\label{clustering_loss_kl}
\end{equation}
where the high confidence (target) assignments $P_{ij}$ can be calculated by first raising $Q_{ij}$ to the second pow and then normalizing by the frequency per cluster as formulated:
\begin{equation}
P_{ij} = \frac{Q_{ij}^2/\sum_i Q_{ij}}{\sum_{j'} Q_{ij'}^2/\sum_i Q_{ij'}}.
\label{Q_calculation}
\end{equation}
Here, the distribution $\mathbf{P}$ is considered the sharpened clustering distribution, which contains more high-confidence clustering information. By aligning the original clustering distribution $\mathbf{Q}$ and sharpened clustering distribution $\mathbf{P}$, the clusters are iteratively refined, therefore improving clustering performance.

\begin{figure}[h]
\centering
\small
\begin{minipage}{0.9\linewidth}
\centerline{\includegraphics[width=1\textwidth]{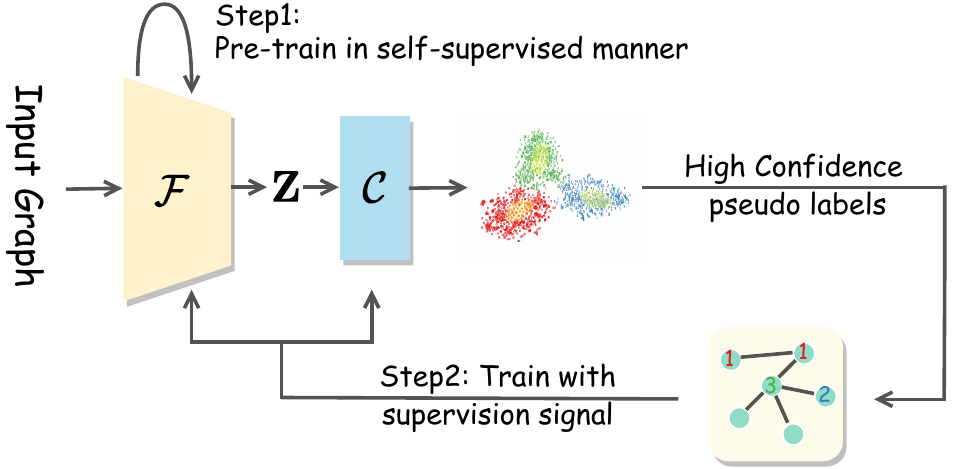}}
\end{minipage}
\caption{The clustering pseudo label technology in the neural clustering methods. In the first step, the encoder $\mathcal{F}$ is trained by the self-supervised pre-text tasks to obtain the discriminative node embeddings $\mathbf{Z}$. Then the clustering method $\mathcal{C}$ is performed on the learned node embeddings. In the second step, high-confidence samples are selected as the pseudo labels. These pseudo labels serve as the supervision signal to further fine-tune the encoder $\mathcal{F}$.}
\label{pseudo_label_figure}
\end{figure}

Similarly, in other two-stage neural clustering methods \cite{GDCL,SGCMC,SCAGC,CDRS,IDCRN}, the high confidence clustering pseudo labels are adopted as the supervision signal for better clustering as shown in Figure \ref{pseudo_label_figure}. For instance, in GDCL, \cite{GDCL} debias the false negative sampling of contrastive learning with the pseudo clustering labels. Besides, \cite{SCAGC} proposes SCAGC to maximize the similarities of intra-cluster nodes while minimizing the similarities of inter-cluster nodes. Moreover, in IDCRN, \cite{IDCRN} utilizes the high confidence pseudo labels to refine the adjacency matrix and guide the learned embeddings to recover the affinity matrix even across views. In addition, \cite{SGCMC} design a cross-entropy based objective function to guide the network to classify the clustering pseudo labels in the SGCMC model. Furthermore, \cite{CDRS} collaborates the pseudo node classification task with clustering and augments a pseudo-label set to further improve the clustering performance. 

% previous methods are 

% In addition, 

Although previous methods achieve promising performance, the scalability of the KL divergence loss and pseudo-label promoting is limited. We analyze the complexity of the KL divergence loss in Eq. \eqref{clustering_loss_kl} as follows. This KL divergence loss function optimizes the clustering distribution with the whole data, therefore leading to high time and space costs. The time complexity and space complexity of calculating the KL divergence loss is $\mathcal{O}(nkd)$ and $\mathcal{O}(nk+nd+kd)$, where $n$ denotes the node number, $k$ denotes the cluster number and $d$ denotes the dimension number of latent features. On the large graphs, the node number $n$ becomes very large, e.g., 111 million nodes on ogbn-papers100M \cite{OGB_dataset}. Therefore the KL divergence loss easily leads to the out-of-memory error or the long running time problem. To alleviate these problems, Dink-Net \cite{dink_net} is proposed by the idea of cluster dilation and cluster shrink. Concretely, the graph encoders are first pre-trained with a discriminative pre-text task. Then the cluster centers are initialized on the learned node embeddings in the $k$-means++ manner \cite{K-means}. Note that these cluster centers are assigned as the tensors with gradient, thus can be optimized with gradient descent algorithms \cite{SGD,ADAM}. Furthermore, the clustering distribution is optimized by minimizing the cluster dilation and cluster shrink loss functions as follows. 

\begin{table}[!t]
\centering
\caption{The statistical information of eleven attribute graph datasets. ``\# Nodes'' denotes the number of nodes, ``\# Edges'' denotes the number of edges, ``\# Feature Dims'' denotes the feature dimension number, and ``\# Classes'' denotes the number of categories. }
\label{statistics_information}
% \scalebox{0.9}{
\resizebox{\linewidth}{!}{
\begin{tabular}{@{}ccccc@{}}
\toprule
\textbf{Dataset}     & \textbf{\# Nodes} & \textbf{\# Edges} & \textbf{\# Feature Dims} & \textbf{\# Classes}  \\ 
\midrule
BAT              & 131             & 81             & 1,038                    & 4                    \\
EAT              & 399             & 203             & 5,994                    & 4                    \\
UAT              & 1,190             & 239             & 13,599                    & 4                    \\
Cora              & 2,708             & 5,278             & 1,433                    & 7                    \\
CoraFull              & 19,793             & 63,421             & 8,710                    & 70                    \\
CiteSeer          & 3,327             & 4,614             & 3,703                    & 6                    \\
ACM          & 3,025             & 1,870             & 13,128                    & 3                    \\
DBLP          & 4,057             & 334             & 3,528                    & 4                    \\
Amazon-Photo              & 7,650             & 119,081           & 745                      & 8                    \\
ogbn-arxiv        & 169,343           & 1,166,243         & 128                      & 40                   \\
Reddit            & 232,965           & 23,213,838        & 602                      & 41                   \\
ogbn-products     & 2,449,029         & 61,859,140        & 100                      & 47                   \\
ogbn-papers100M   & 111,059,956       & 1,615,685,872     & 128                      & 172                  \\
\bottomrule
\end{tabular}}
\end{table}

\begin{equation} 
\begin{aligned}
\mathcal{L}_{\textrm{dilation}} &= \\  &\frac{-1}{(k-1)k} \sum_{i=1}^{k} \sum_{j=1, j\neq i}^{k} \left\|\mathbf{c}_i-\mathbf{c}_j\right\|^2,
\end{aligned}
\label{eq:dilation_loss}
\end{equation}
\begin{equation} 
\mathcal{L}_{\textrm{shrink}} =  \frac{1}{bk} \sum_{i=1}^{b}\sum_{j=1}^{k}\left\|\mathbf{z}_i-\mathbf{c}_j\right\|^2,
\label{eq:shrink_loss}
\end{equation}
\begin{equation} 
\mathcal{L}_{\text{Dink-Net}} = \mathcal{L}_{\textrm{dilation}} + \mathcal{L}_{\textrm{shrink}},
\label{eq:dink_net_loss}
\end{equation}
where the first term $\mathcal{L}_{\text{dilation}}$ denotes the cluster dilation loss, and the second term $\mathcal{L}_{\text{shrink}}$ denotes the cluster shrink loss. Besides, $\mathbf{Z} \in \mathbb{R}^{n \times d}, \mathbf{C} \in \mathbb{R}^{k, d}, n, k, b, d$ denotes the node embeddings, cluster center embeddings, node number, cluster number, and hidden feature dimension. In Eq. \ref{eq:dilation_loss}, the cluster dilation loss aims to learn the network to push away different cluster centers. In addition, in Eq. \ref{eq:shrink_loss}, the cluster shrink loss aims to pull the samples to the cluster centers. Note that the cluster shrink loss pulls batch samples to the cluster centers and easily scales to large graphs. Besides, each sample will be pulled to all cluster centers rather than the corresponding nearest cluster centers. The authors claim this design alleviates the confirmation bias problem \cite{confirmation_bias}. The time complexity and space complexity of calculating $\mathcal{L}_{\text{Dink-Net}}$ is $\mathcal{O}(bkd+k^2d+bd)$ and $\mathcal{O}(bk+bd+kd)$, respectively.

% The cluster dilation loss aims to push away the 

% When the node number in graphs become large, such as 111 million 

% loss function optimizes the clustering distribution with the whole data. Thus, the calculation and optimization process is complex and resource-consuming, leading to $\mathcal{O}(NKd)$ time complexity and $\mathcal{O}(NK+Nd+Kd)$ space complexity.

Although verified effective, the two-stage neural clustering methods heavily depend on the excellent initialization of cluster centers. And the precise cluster center initialization relies on the promising network pre-training and the auxiliary of traditional clustering methods like $k$-Means, leading to high training and deployment costs. To overcome this shortcoming, many one-stage methods \cite{SCGC,AGC-DRR,GCC} have been gradually proposed recently. For example, \cite{SCGC} proposes a new method termed SCGC with Laplacian smoothing and neighborhood-orient contrastive learning. Besides, in AGC-DRR \cite{AGC-DRR}, the clustering sub-network is developed to directly predict the probabilities of cluster-ID for each sample. Moreover, rather than dealing with clustering as a downstream task, \cite{GCC} proposes a unified framework termed GCC by minimizing the difference between node embeddings and reconstructed cluster embeddings. The one-stage neural clustering will be a promising future direction. In the next section, we analyze and summarize the technical challenges and opportunities in the deep graph clustering field in detail.

\begin{table*}[!t]
\centering
\caption{Graph data quality testing experiments. Clustering performance of HSAN \cite{HSAN} on five datasets with missing information including dropped node attributes and dropped edges. Experimental results are the mean value of ten runs. }
\label{drop}
% \scalebox{1.0}{
\resizebox{0.80\linewidth}{!}{
\begin{tabular}{ccc|ccccc|ccccc}
\hline
                          &     & \multirow{2}{*}{\textbf{Origin}} & \multicolumn{5}{c|}{\textbf{Attribute Drop Rate}}  & \multicolumn{5}{c}{\textbf{Edge Drop Rate}}          \\ \cline{4-13} 
                          &     &                         & 10\%   & 30\%   & 50\%   & 70\%   & 90\%   & 10\%   & 30\%   & 50\%   & 70\%   & 90\%    \\ \hline
\multirow{4}{*}{\textbf{Cora}}     & \textbf{ACC} & 77.73                   & 75.22 & 74.00 & 68.07 & 63.74 & 46.16 & 77.13 & 75.00 & 71.53 & 62.96 & 48.81 \\
                          & \textbf{NMI} & 59.45                   & 57.93 & 54.73 & 48.26 & 40.62 & 21.21 & 59.55 & 56.74 & 51.69 & 42.54 & 29.54 \\
                          & \textbf{ARI} & 57.62                   & 54.76 & 51.72 & 42.05 & 36.01 & 16.42 & 58.03 & 53.81 & 47.31 & 38.10 & 18.98 \\
                          & \textbf{F1}  & 76.16                   & 73.70 & 72.78 & 67.72 & 62.91 & 45.39 & 74.97 & 72.09 & 68.48 & 59.34 & 47.12 \\ \hline
\multirow{4}{*}{\textbf{CiteSeer}} & \textbf{ACC} & 71.36                   & 69.64 & 64.52 & 62.26 & 51.21 & 36.13 & 70.83 & 67.66 & 62.48 & 53.64 & 55.06 \\
                          & \textbf{NMI} & 45.03                   & 42.01 & 35.56 & 32.33 & 19.83 & 8.66  & 44.53 & 42.12 & 35.30 & 26.80 & 27.50 \\
                          & \textbf{ARI} & 46.81                   & 43.94 & 36.19 & 32.23 & 19.74 & 8.02  & 45.95 & 38.75 & 31.38 & 22.38 & 23.50 \\
                          & \textbf{F1}  & 63.63                   & 62.25 & 56.98 & 54.09 & 45.08 & 28.75 & 63.44 & 59.53 & 56.06 & 49.59 & 48.84 \\ \hline
\multirow{4}{*}{\textbf{BAT}}      & \textbf{ACC} & 77.86                   & 77.61 & 75.32 & 74.30 & 76.08 & 70.74 & 65.90 & 56.74 & 49.87 & 48.35 & 51.15 \\
                          & \textbf{NMI} & 53.00                   & 53.34 & 50.19 & 49.40 & 50.96 & 45.29 & 41.70 & 30.36 & 22.33 & 24.82 & 22.75 \\
                          & \textbf{ARI} & 51.16                   & 51.26 & 47.42 & 46.25 & 48.61 & 42.26 & 36.32 & 23.94 & 15.40 & 15.42 & 17.09 \\
                          & \textbf{F1}  & 77.80                   & 77.48 & 75.15 & 74.02 & 75.85 & 69.40 & 62.71 & 53.10 & 45.43 & 43.00 & 50.50 \\ \hline
\multirow{4}{*}{\textbf{EAT}}      & \textbf{ACC} & 57.73                   & 57.73 & 58.15 & 58.06 & 57.39 & 55.97 & 55.22 & 52.88 & 47.70 & 44.78 & 45.36 \\
                          & \textbf{NMI} & 34.12                   & 34.35 & 34.86 & 33.83 & 33.00 & 31.73 & 33.10 & 29.38 & 22.26 & 16.79 & 17.88 \\
                          & \textbf{ARI} & 27.22                   & 27.37 & 28.17 & 27.33 & 26.99 & 25.58 & 26.16 & 22.89 & 15.57 & 11.40 & 13.69 \\
                          & \textbf{F1}  & 58.09                   & 57.98 & 58.31 & 58.21 & 57.20 & 55.85 & 51.02 & 47.87 & 45.79 & 44.43 & 43.42 \\ \hline
\multirow{4}{*}{\textbf{UAT}}      & \textbf{ACC} & 56.16                   & 56.89 & 55.18 & 55.29 & 54.45 & 55.77 & 52.55 & 46.69 & 42.86 & 42.94 & 47.90 \\
                          & \textbf{NMI} & 26.89                   & 26.60 & 25.51 & 23.55 & 23.77 & 24.68 & 20.30 & 14.58 & 13.57 & 15.04 & 19.06 \\
                          & \textbf{ARI} & 25.21                   & 24.49 & 23.11 & 23.23 & 22.77 & 24.79 & 19.40 & 13.82 & 09.63 & 11.98 & 16.28 \\
                          & \textbf{F1}  & 55.99                   & 54.33 & 51.46 & 53.86 & 51.91 & 55.05 & 49.60 & 43.53 & 40.65 & 38.13 & 44.13 \\ \hline
\end{tabular}}
\end{table*}

% Time and space analyses of deep graph clustering methods. The experimental costs are obtained on the Cora dataset. ``-'' means ruining on CPU.

% % VIS t-SNE
% \begin{figure*}[!t]
% \centering
% \small
% \begin{minipage}{0.139\linewidth}
% \centerline{\includegraphics[width=\textwidth]{vis-attribute_0.png}}
% \vspace{2pt}
% \centerline{Raw Attribute}
% \end{minipage}
% \begin{minipage}{0.139\linewidth}
% \centerline{\includegraphics[width=\textwidth]{vis-cora_DEC.png}}
% \vspace{2pt}
% \centerline{DEC}
% \end{minipage}
% \begin{minipage}{0.139\linewidth}
% \centerline{\includegraphics[width=\textwidth]{vis-cora_MVGRL.png}}
% \vspace{2pt}
% \centerline{MVGRL}
% \end{minipage}
% \begin{minipage}{0.139\linewidth}
% \centerline{\includegraphics[width=\textwidth]{vis-cora_mincutpool.png}}
% \vspace{2pt}
% \centerline{MinCutPool}
% \end{minipage}
% \begin{minipage}{0.139\linewidth}
% \centerline{\includegraphics[width=\textwidth]{vis-cora_GRACE.png}}
% \vspace{2pt}
% \centerline{DCRN}
% \end{minipage}
% \begin{minipage}{0.139\linewidth}
% \centerline{\includegraphics[width=\textwidth]{vis-cora_S3GC.png}}
% \vspace{2pt}
% \centerline{S$^3$GC}
% \end{minipage}
% \begin{minipage}{0.139\linewidth}
% \centerline{\includegraphics[width=\textwidth]{vis-cora_ours.png}}
% \vspace{2pt}
% \centerline{Ours}
% \end{minipage}
% % \setlength{\belowcaptionskip}{-10pt}
% \caption{\textit{t}-SNE visualization of seven methods on the Cora dataset.}
% \label{fig:visualization}  
% \end{figure*}

\begin{table*}[!t]
\centering
\caption{Graph data quality testing experiments. Clustering performance of HSAN \cite{HSAN} on five datasets with noisy information including noised node attributes and noised edges. Experimental results are the mean value of ten runs. }
\label{noise}
\resizebox{0.80\linewidth}{!}{
\begin{tabular}{ccc|cccc|cccc}
\hline
                          &     & \multirow{2}{*}{\textbf{Origin}} & \multicolumn{4}{c|}{\textbf{Standard Deviation of Attribute Noises}} & \multicolumn{4}{c}{\textbf{Standard Deviation of Edge Noises}} \\ \cline{4-11} 
                          &     &                         & 0.01    & 0.1     & 1      & 10     & 0.01  & 0.1   & 1     & 10    \\ \hline
\multirow{4}{*}{\textbf{Cora}}     & \textbf{ACC} & 77.73                   & 77.31   & 76.53   & 46.82  & 36.15  & 37.20 & 21.25 & 30.24 & 30.24 \\
                          & \textbf{NMI} & 59.45                   & 59.04   & 58.37   & 22.18  & 13.53  & 11.16 & 0.51  & 0.40  & 0.40  \\
                          & \textbf{ARI} & 57.62                   & 59.16   & 56.43   & 17.44  & 8.95,  & 8.53  & 0.48  & 0.02 & 0.02 \\
                          & \textbf{F1}  & 76.16                   & 74.76   & 74.76   & 46.34  & 37.19  & 34.4  & 15.61 & 6.84  & 6.84  \\ \hline
\multirow{4}{*}{\textbf{CiteSeer}} & \textbf{ACC} & 71.36                   & 70.39   & 68.79   & 37.47  & 27.35  & 28.27 & 21.57 & 21.07 & 21.07 \\
                          & \textbf{NMI} & 45.03                   & 44.54   & 41.85   & 8.44   & 3.52   & 3.70  & 0.35  & 0.29  & 0.29  \\
                          & \textbf{ARI} & 46.81                   & 46.16   & 43.64   & 8.06   & 1.55   & 2.21  & 0.20  & 0.00  & 0.00  \\
                          & \textbf{F1}  & 63.63                   & 64.12   & 62.40   & 30.49  & 17.70  & 21.85 & 15.53 & 5.91  & 5.91  \\ \hline
\multirow{4}{*}{\textbf{BAT}}      & \textbf{ACC} & 77.86                   & 78.37   & 78.37   & 66.41  & 58.52  & 66.41 & 52.67 & 27.48 & 27.48 \\
                          & \textbf{NMI} & 53.00                   & 54.15   & 53.07   & 46.50  & 40.53  & 38.86 & 23.62 & 4.36  & 4.36  \\
                          & \textbf{ARI} & 51.16                   & 52.27   & 52.33   & 42.14  & 31.70  & 34.02 & 18.13 & 0.15, & 0.15  \\
                          & \textbf{F1}  & 77.80                   & 78.30   & 78.19   & 62.76  & 56.32  & 66.12 & 52.45 & 12.25 & 12.25 \\ \hline
\multirow{4}{*}{\textbf{EAT}}      & \textbf{ACC} & 57.73                   & 57.98   & 56.64   & 53.22  & 50.71  & 55.81 & 50.71 & 26.07 & 26.07 \\
                          & \textbf{NMI} & 34.12                   & 34.36   & 33.59   & 32.13  & 32.17  & 31.58 & 23.27 & 1.46  & 1.46  \\
                          & \textbf{ARI} & 27.22                   & 28.06   & 25.92   & 25.64  & 23.73  & 24.07 & 17.30 & 0.01  & 0.01  \\
                          & \textbf{F1}  & 58.09                   & 58.11   & 57.12   & 48.54  & 45.92  & 56.37 & 51.62 & 11.24 & 11.24 \\ \hline
\multirow{4}{*}{\textbf{UAT}}      & \textbf{ACC} & 56.16                   & 55.57   & 56.97   & 53.39  & 54.17  & 47.00 & 40.25 & 25.38 & 25.38 \\
                          & \textbf{NMI} & 26.89                   & 25.59   & 26.23   & 24.33  & 24.88  & 15.72 & 9.26  & 0.50  & 0.50  \\
                          & \textbf{ARI} & 25.21                   & 25.41   & 25.60   & 22.02  & 23.88  & 15.41 & 9.10  & 0.00  & 0.00  \\
                          & \textbf{F1}  & 55.99                   & 54.86   & 53.64   & 47.29  & 52.79  & 45.42 & 36.97 & 10.56 & 10.56 \\ \hline
\end{tabular}}
\end{table*}

\section{Challenge \& Opportunity}
\label{challenge_opportunity}
In recent years, we have witnessed the fast growth of deep graph clustering. More and more methods have been proposed and achieved promising performance. However, most methods are under some perfect assumption, and there are still many challenges in the field of deep graph clustering. From this motivation, this section aims to summarize the main technical challenges. As shown in Figure \ref{challenges_figure}, the main challenges in deep graph clustering contain five aspects, i.e., graph data quality, stability, scalability, discriminative capability, and unknown cluster number. In the following five sub-sections, we will analyze these challenges via experiments and provide potential solutions and opportunities.

% in this section, we first summarize and analyze the urge challenges in deep graph clustering. Then, we discuss the corresponding potential solutions and opportunities. 

% In this section, we first summarize and analyze the urge challenges in deep graph clustering. Subsequently, the corresponding potential solutions and opportunities are discussed. 

\begin{table*}[!t]
    \centering
    \caption{Stability testing of deep graph clustering methods. Average clustering performance of eleven methods on six datasets. The experimental results are obtained with ten runs and presented with standard deviation.}
    \small
    % \scalebox{0.90}{ 
    \resizebox{0.80\linewidth}{!}{
    % \resizebox{\linewidth}{!}{
        \begin{tabular}{@{}ccccccccccccc@{}}
            \toprule
            \textbf{Dataset}           & \textbf{Metric} & \textbf{$k$-Means} & \textbf{AE} & \textbf{DEC} & \textbf{IDEC} & \textbf{GAE} & \textbf{DAEGC}         & \textbf{ARGA} & \textbf{SDCN} & \textbf{MVGRL}         & \textbf{DFCN}          & {\color[HTML]{000000} \textbf{DCRN}} \\ \midrule
                                       & \textbf{ACC}             & 0.65       & 0.35  & 0.56   & 0.62    & 1.22   & 0.48             & 0.59    & 1.81    & 1.02             & {0.80} & {0.25}                  \\
                                       & \textbf{NMI}             & 0.38       & 0.16  & 0.28   & 0.50    & 0.91   & 0.45             & 0.92    & 1.34    & 0.63             & {1.00} & {0.44}                  \\
                                       & \textbf{ARI}             & 0.39        & 0.43  & 0.39   & 0.60    & 1.40   & 0.52             & 0.91    & 2.01    & 0.50              & {1.50} & {0.46}                  \\
            \multirow{-4}{*}{\textbf{DBLP}}     & \textbf{F1}              & 0.27       & 0.36  & 0.51   & 0.56    & 2.23   & 0.67             & 0.66    & 1.51    & 1.51             & {0.80} & {0.26}                  \\ \midrule
                                       & \textbf{ACC}             & 3.17       & 0.13  & 0.20   & 1.42    & 0.80   & 1.39             & 0.49    & 0.31    & 0.36             & {0.20} & {0.18}                  \\
                                       & \textbf{NMI}             & 3.22       & 0.08  & 0.30   & 2.40    & 0.65   & 0.86             & 0.71    & 0.32    & 0.40             & {0.20} & {0.35}                  \\
                                       & \textbf{ARI}             & 3.02       & 0.14  & 0.36   & 2.65    & 1.18   & 1.24             & 0.70    & 0.43    & 0.73             & {0.30} & {0.30}                  \\
            \multirow{-4}{*}{\textbf{CiteSeer}}     & \textbf{F1}              & 3.53       & 0.11  & 0.17   & 1.39    & 0.82   & 1.32             & 0.31    & 0.24    & 0.39             & {0.20} & {0.21}                  \\ \midrule
                                       & \textbf{ACC}             & 0.71       & 0.08  & 0.76   & 0.52    & 1.44   & 2.83             & 0.36    & 0.18    & 0.76             & {0.20} & {0.20}                  \\
                                       & \textbf{NMI}             & 0.46       & 0.16  & 1.51   & 1.16    & 1.92   & 4.15             & 0.82    & 0.25    & 1.40             & {0.40} & {0.61}                  \\
                                       & \textbf{ARI}             & 0.69       & 0.16  & 1.87   & 1.50    & 3.10   & 3.89             & 0.86    & 0.40    & 1.76             & {0.40} & {0.52}                  \\
            \multirow{-4}{*}{\textbf{ACM}}      & \textbf{F1}              & 0.74       & 0.08  & 0.74   & 0.48    & 1.33   & 2.79             & 0.35    & 0.19    & 0.72             & {0.20} & {0.20}                  \\ \midrule
                                       & \textbf{ACC}             & 0.76       & 0.08  & 0.08   & 0.08    & 2.48   & 0.01             & 2.30    & 0.81    & 1.79             & {0.80} & {0.13}                  \\
                                       & \textbf{NMI}             & 1.33       & 0.30  & 0.05   & 0.08    & 2.79   & 0.03             & 2.76    & 0.83    & 1.31             & {1.00} & {0.24}                  \\
                                       & \textbf{ARI}             & 0.44        & 0.47  & 0.04   & 0.07    & 4.57   & {0.02} & 4.41    & 1.23    & 0.47             & 0.84             & {0.20}                  \\
            \multirow{-4}{*}{\textbf{Amazon-Photo}}     & \textbf{F1}              & 0.51       & 0.20  & 0.12   & 0.11    & 1.76   & 0.02             & 1.95    & 1.49    & 2.39             & {0.31} & {0.12}                  \\ \midrule
                                       & \textbf{ACC}             & 0.01       & 0.31  & 0.09   & 0.34    & 0.81   & 0.03             & 0.12    & 1.30    & 0.52             & {0.07} & {0.07}                  \\
                                       & \textbf{NMI}             & 0.02       & 0.57  & 0.14   & 0.29    & 3.54   & 0.03             & 0.17    & 2.04    & {1.45} & 0.13             & {0.08}                  \\
                                       & \textbf{ARI}             & 0.01       & 0.67  & 0.13   & 0.39    & 1.39   & 0.04             & 0.17    & 2.07    & 1.06             & {0.11} & {0.12}                  \\
            \multirow{-4}{*}{\textbf{PubMed}}   & \textbf{F1}              & 0.01       & 0.29  & 0.10   & 0.32    & 0.85   & {0.02} & 0.13    & 1.21    & 0.36             & 0.07             & {0.08}                  \\ \midrule
                                       & \textbf{ACC}             & 1.10       & 0.19  & 0.45   & 0.31    & 0.81   & 1.00             & 0.43    & 0.40    & 2.95             & {0.81} & {0.60}                  \\
                                       & \textbf{NMI}             & 0.84       & 0.25  & 0.24   & 0.28    & 0.75   & 0.73             & 0.25    & 0.39    & 3.95             & {0.41} & {0.35}                  \\
                                       & \textbf{ARI}             & 0.57        & 0.27  & 0.29   & 0.22    & 0.86   & 0.47             & 0.24    & 0.27    & 2.63             & {0.48} & {0.49}                  \\
            \multirow{-4}{*}{\textbf{CoraFULL}} & \textbf{F1}              & 1.09       & 0.30  & 0.58   & 0.41    & 0.75   & 1.33             & 0.41    & 0.43    & 2.87             & {0.87} & {0.76}                  \\ \bottomrule
        \end{tabular}
    }
    \label{stability}
\end{table*}

\subsection{Graph Data Quality}
The existing conventional deep graph clustering methods always are under the assumption that the input graphs are perfectly correct, namely, the connections between nodes are complete and correct and the node information is also complete and precise. However, these assumptions do not always hold, especially in the industrial scenario. The real-world graph data are always of low quality. The noises come from two aspects, i.e., nodes and edges. Firstly, for the nodes in graphs, the node attributes may contain error information or incomplete information. To be specific, for the error information, they are caused by error records and easily spreads in the whole graph during the message-passing process. In addition, incomplete information contains two types, i.e., partly attribute missing and completely attribute missing. The latter one is more difficult to process. Secondly, for the edges in graphs, they may also contain the error information and incomplete information. Concretely, the error information denotes the error connections between nodes, while the incomplete information denotes the missing edges, which should exist in the correct graph. Besides, the conventional graph convolutional network encoders \cite{GCN,graphsage} are based on the homophily assumption \cite{homo_assume}, i.e., the connected nodes contain similar semantics. However, in the real-world scenario, the connected nodes may not have similar features. For example, in the heterophily graphs \cite{heterophily_survey} such as fraudulent networks, the hackers and regular users will construct dense connections, but they do not share similar underlying semantics.

To verify this challenge in deep graph clustering, we conduct experiments on one of the representative state-of-the-art methods, HSAN \cite{HSAN}, on five datasets, including Cora, CiteSeer, BAT, EAT, and UAT. The detailed information of the datasets can be found in Table \ref{statistics_information}. The experimental results are demonstrated in Table \ref{drop} and Table \ref{noise}.

In Table \ref{drop}, we evaluate the clustering performance of HSAN \cite{HSAN} with missing data. The missing information contains node attributes and edges. And the drop rates are set to 10\%, 30\%, 50\%, 70\%, and 90\%, respectively. From the experimental results, three conclusions are drawn. Firstly, when the information on graphs is missing, the deep graph clustering method can not achieve promising performance. Secondly, the clustering performance drops when the drop rate increases. Thirdly, the node attributes are more critical to the clustering performance on some citation graphs, such as the CiteSeer dataset, while the edges are more critical to the clustering performance on some airport activity graphs, such as the BAT dataset.

In addition, in Table \ref{noise}, the performance of HSAN \cite{HSAN} with noisy information is tested. Similarly, the noised information contains the noised node attribute and noised edges. Concretely, for the node attributes, we add the Gaussian noises to them, and the standard deviations are set to 0.01, 0.1, 1, and 10, respectively. Besides, for the edges in the graph, we add the Gaussian noises to the adjacency matrix, and the standard deviations are set to 0.01, 0.1, 1, and 10, respectively. We have two conclusions from the experimental results. Firstly, the noises in attributes and edges limit the clustering performance. Secondly, when the noise rate increases, the performance drops significantly.

% \usepackage{multirow}

% noise: attribute noise, random add edge, 

% incomplete: Edge drop, attribute drop, 

% node attributes  

% the node attributes may contain error information or incomplete information. The error information 

% Firstly, the graph data may contain many noises on both the node attributes and the graph structures,

% e.g., the wrong node attributes and error connections between nodes. Secondly, there might be lots of missing information in the graph data, e.g., missing part node attributes, missing connections, or even missing whole node attributes. 

Next, we discuss the potential solutions. Firstly, for the incomplete node attributes or edges, the imputation networks \cite{missing_1,missing_2} can help alleviate the information missing issue. Besides, for the error information, the denoising technologies \cite{denoise_1,denoise_2,denoise_3} might effectively remove the noises in the node attributes and edges. Additionally, to solve the homophily and heterophily problems, researchers propose various methods \cite{heterophily_survey,DGCN}.

\subsection{Stability}
The stability of deep graph clustering algorithms is essential, especially in sensitive domains such as financial risk control, social network anomaly detection, etc. Different from the supervised methods or the semi-supervised methods, deep graph clustering methods group the nodes into disjoint clusters in a purely unsupervised manner, i.e., without any human annotations. Therefore, without the guidance of ground truth, the stability of deep graph clustering models is relatively weak. For example, for the classical clustering method such as $k$-Means \cite{K-means}, its promising performance highly relies on the quality of the initial cluster centers. Similarly, the performance of deep clustering techniques \cite{DEC,IDEC} is also sensitive to the initialization of the neural networks and trainable cluster centers. Next, we systematically analyze the randomness in the deep graph clustering methods. It mainly consists of two parts as follows. Firstly, one of the key factors in achieving promising clustering performance is the excellent representation capability of the learned node embeddings. Thus, the initialization and training processes of the embedding networks influence the stability of deep graph clustering methods. Secondly, a clustering method with strong discriminative capability is another critical factor. Therefore, initialization and optimization processes of the neural cluster centers also easily affect the stability of deep graph clustering methods.

We conduct experiments to test the stability of existing deep graph clustering methods on six datasets, including DBLP, CiteSeer, ACM, Amazon-Photo, PubMed, and CoraFULL. The compared methods include DCRN \cite{DCRN}, DFCN \cite{DFCN}, MVGRL \cite{MVGRL}, DAEGC \cite{DAEGC}, , SDCN \cite{SDCN}, ARGA \cite{ARGA}, GAE \cite{GAE}, DEC \cite{DEC}, IDEC \cite{IDEC}, AE \cite{AE}, $k$-Means \cite{K-means}. All experimental results are obtained with ten runs. And the results are reported with standard deviation. The experimental results indicate that the stability of recent methods seems good. However, there are various problems with their implementation. Firstly, the classical method $k$-Means initializes the cluster centers various times to find excellent initialization. Secondly, most of the existing methods such as DEC \cite{DEC}, IDEC \cite{IDEC}, DAEGC \cite{DAEGC}, ARGA \cite{ARGA}, SDCN \cite{SDCN}, MVGRL \cite{MVGRL}, DFCN \cite{DFCN}, and DCRN \cite{DCRN} rely on the well pre-trained node embeddings and cluster center embeddings. And the manual trial processes of network pre-training and center initialization seem to increase the stability of deep graph clustering methods. However, in the real-world scenario, implementing these procedures is costly. Thus, we consider that the stability of the existing deep graph clustering methods is overestimated. It will be a promising direction to improve the robustness and stability of deep graph clustering methods.

In the existing studies, researchers conduct ten runs with different random seeds to alleviate the randomness influence on the experimental results. We consider that some optimization strategies \cite{stable_1,stable_2} might enhance the stability of deep graph clustering.

% Thus, the existing works always fix the random seed and run algorithms ten times to alleviate the influence of randomness. 

% experiments: different randomness
% 1) Train the encoding network on the large graph. 2) Group the large number of nodes in the graph into separate clusters. 

\begin{table*}[!t]
\centering
\caption{Scalability testing experiments on four large graphs. ``OOM'' indicates that method raises the out-of-memory failure.}
\label{compare_table}
\setlength{\tabcolsep}{4pt}
% \fontsize{8}{10}\selectfont 
% \resizebox{\linewidth}{!}{
% \scalebox{0.83}{
\resizebox{\linewidth}{!}{
\begin{tabular}{@{}ccccccccccccccccc@{}}
\toprule
\multirow{2}{*}{\textbf{Dataset}}         & \multirow{2}{*}{\textbf{Metric}} & \multirow{2}{*}{\textbf{$k$-Means}} & \multirow{2}{*}{\textbf{DEC}} & \multirow{2}{*}{\textbf{DCN}} & \multirow{2}{*}{\textbf{node2vec}} & \multirow{2}{*}{\textbf{DGI}} & \multirow{2}{*}{\textbf{AGE}}  & \multirow{2}{*}{\textbf{MVGRL}} & \multirow{2}{*}{\textbf{BGRL}} & \multirow{2}{*}{\textbf{GRACE}} & \multirow{2}{*}{\textbf{ProGCL}} & \multirow{2}{*}{\textbf{AGC-DRR}} & \multirow{2}{*}{\textbf{DCRN}} & \multirow{2}{*}{\textbf{S$^3$GC}} & \multirow{2}{*}{\textbf{Dink-Net}} \\
&                                  &                                  &                               &                               &                                    &                               &                               &                                      &                                 &                                &                                 &                                  &                                   &                                &                                &                                \\ \midrule
% % \midrule
\multirow{4}{*}{\textbf{ogbn-arXiv}}      & \textbf{ACC}                     & 18.11                            & 21.25                         & 19.91                         & 29.00                              & 31.40                         & \multirow{4}{*}{OOM}                                    & \multirow{4}{*}{OOM}            & 22.70                          & \multirow{4}{*}{OOM}            & 29.86                            & \multirow{4}{*}{OOM}              & \multirow{4}{*}{OOM}           & {35.00}                    & 43.68                 \\
& \textbf{NMI}                     & 22.13                            & 25.14                         & 23.81                         & 40.60                              & 41.20                         &                                                        &                                 & 32.10                          &                                 & 37.51                            &                                   &                                & 46.30                & 43.73                    \\
& \textbf{ARI}                     & 7.43                             & 10.28                         & 8.25                          & 19.00                              & 22.30                         &                                                            &                                 & 13.00                          &                                 & 25.74                            &                                   &                                &  27.00                   & 35.22                \\
& \textbf{F1}                      & 12.94                            & 15.57                         & 13.06                         & 22.00                              & 23.00                         &                                                       &                                 & 16.60                          &                                 & 21.79                            &                                   &                                & {23.00}                    & 26.92              \\ \midrule
% % \midrule
\multirow{4}{*}{\makecell[c]{\textbf{ogbn-} \\ \textbf{products}}}   & \textbf{ACC}                     & 18.11                            & 23.79                         & 24.50                         & 35.70                              & 32.00                         & \multirow{4}{*}{OOM}          & \multirow{4}{*}{OOM}            & \multirow{4}{*}{OOM}           & \multirow{4}{*}{OOM}            & 35.21                            & \multirow{4}{*}{OOM}              & \multirow{4}{*}{OOM}           & 40.20                  & 41.09                \\
& \textbf{NMI}                     & 22.13                            & 24.47                         & 21.92                         & 48.90                              & 46.70                         &                                                            &                                 &                                &                                 & 46.59                            &                                   &                                & 53.60                & 50.78                   \\
& \textbf{ARI}                     & 7.43                             & 9.05                          & 10.96                         & 17.00                              & 17.40                         &                                                      &                                 &                                &                                 & 19.87                            &                                   &                                & 23.00                 & 21.08                  \\
& \textbf{F1}                      & 12.94                            & 13.54                         & 13.95                         & 24.70                              & 19.20                         &                                                     &                                 &                                &                                 & 21.55                            &                                   &                                & {25.00}                    & 25.15                \\ \midrule
% % \midrule
\multirow{4}{*}{\textbf{Reddit}}          & \textbf{ACC}                     & 8.90                             & \multirow{4}{*}{OOM}          & \multirow{4}{*}{OOM}          & 70.90                              & 22.40                         & \multirow{4}{*}{OOM}                            & \multirow{4}{*}{OOM}            & \multirow{4}{*}{OOM}           & \multirow{4}{*}{OOM}            & 65.41                            & \multirow{4}{*}{OOM}              & \multirow{4}{*}{OOM}           & 73.60                  & 76.03                 \\
& \textbf{NMI}                     & 11.40                            &                               &                               & 79.20                              & 30.60                         &                               &                                      &                                 &                                                               & 70.48                            &                                   &                                & 80.70                 & 78.91                    \\
& \textbf{ARI}                     & 2.90                             &                               &                               & 64.00                              & 17.00                         &                               &                                      &                                 &                                &                                 63.42                            &                                   &                                & 74.50                & 71.34                   \\
& \textbf{F1}                      & 6.80                             &                               &                               & 55.10                              & 18.30                         &                               &                                      &                                 &                                &                                 51.45                            &                                   &                                & 56.00                   & 67.95                \\ \midrule
% \midrule
\multirow{4}{*}{\makecell[c]{\textbf{ogbn-} \\ \textbf{papers100M}}} & \textbf{ACC}                     & 14.60                            & \multirow{4}{*}{OOM}          & \multirow{4}{*}{OOM}          & { 17.50}                        & 15.10                         & \multirow{4}{*}{OOM}          & \multirow{4}{*}{OOM}                 & \multirow{4}{*}{OOM}                     & \multirow{4}{*}{OOM}            & \multirow{4}{*}{OOM}             & \multirow{4}{*}{OOM}              & \multirow{4}{*}{OOM}           & 17.30                          & 26.67                 \\
& \textbf{NMI}                     & 37.33                            &                               &                               & 38.00                              & 41.60                         &                               &                                      &                                 &                                &                                 &                                  &                                   &                                45.30                   & 54.92                 \\
& \textbf{ARI}                     & 7.54                             &                               &                               & 11.20                        & 9.60                          &                               &                                      &                                 &                                &                                 &                                  &                                   &                                11.00                          & 18.01                \\
& \textbf{F1}                      & 10.45                            &                               &                               & 11.10                              & 11.10                         &                               &                                      &                                 &                                &                                 &                                  &                                   &                                11.80                   & 19.48                \\ 
% \bottomrule
\end{tabular}}
\end{table*}

\begin{table*}[!t]
\centering
\caption{Efficiency analyses of eight deep graph clustering methods. The left two columns denote the time and space complexity analyses. The right two columns denote the GPU and time costs. Experimental results obtained on the Cora dataset. $n$, $b$, $k$, $l$, $g$, $d'$, and $d$ denote node number, batch size, cluster number, edge number, average degree of graph, attribute dimension, and latent feature dimension.}
\label{table:complexity}
% \scalebox{1.1}{
\resizebox{0.8\linewidth}{!}{
\begin{tabular}{@{}ccccc@{}}
\toprule
\textbf{Method}     & \textbf{Time Complexity (per iteration)} & \textbf{Space Complexity} & \textbf{GPU Memory Cost (MB)} & \textbf{Time Cost (s)} \\ \midrule
\textbf{DEC}            &        $\mathcal{O}(nkd)$                  &             $\mathcal{O}(nk+nd+kd)$              &          1294            &       14.59             \\
\textbf{node2vec}            &        $\mathcal{O}(bd)$                  &             $\mathcal{O}(nd)$              & -                     &         111.03           \\
\textbf{DGI}                 &        $\mathcal{O}(ld'+nd^2)$                    &         $\mathcal{O}(l+nd+d^2)$                   &            3798          &               19.03     \\
\textbf{MVGRL}               &  $\mathcal{O}(n^2d+nd^2)$                        &          $\mathcal{O}(n^2+nd+d^2)$                 &                  9466    &           168.20         \\
\textbf{GRACE}               &     $\mathcal{O}(n^2d+ld+d^2)$                      &         $\mathcal{O}(l+nd)$                  &                 1292     &          44.77          \\
\textbf{BGRL}                &        $\mathcal{O}(ld+nd^2)$                  &         $\mathcal{O}(l+nd+d^2)$                  &                1258      &       44.18             \\
\textbf{S$^3$GC}                &         $\mathcal{O}(ngd^2)$                 &              $\mathcal{O}(nd+bgd+d^2)$             &                   1474   &      508.21              \\ 
\textbf{Dink-Net}                &         $\mathcal{O}(nkd+k^2d+bd)$                 &              $\mathcal{O}(bk+bd+kd)$             &             1248         &            35.09        \\ \bottomrule
\end{tabular}}
\end{table*}

\subsection{Scalability}
In this section, we discuss the scalability of deep graph clustering methods. Although achieving promising performance, most previous deep graph clustering methods can not scale to large-scale graph datasets, such as ogbn-papers100M \cite{OGB_dataset}. The heavy time and memory costs mainly come from two parts, as follows. Firstly, the graph encoders need to be trained on large-scale graphs. Secondly, the clustering algorithms need to group a large number of nodes into separate clusters. 

For the first part, batch training \cite{mini-batch_training} will be a good solution. Concretely, researchers can partition the large graph into a mini-batch graph via sub-graph extraction technology like random walk \cite{random_walk} and then train the networks on the sub-graphs. Here, unlike images or texts, one of the essential characteristics of graph data is the relational connections between nodes. Thus, keeping the original connection information after the graph partition is vital for the model's performance.

For the second part, given the learned node embeddings $\mathbf{Z} \in \mathbb{R}^{n \times d}$, the classical $k$-Means clustering algorithm will take $\mathcal{O}(tknd)$ time complexity and $\mathcal{O}(nd+kd)$ space complexity. Here, $n$, $k$, and $d$ are the node number, cluster number, and feature dimension number, respectively. Besides, $t$ is the iteration number of $k$-Means. On the large graph, the node number $n$ is very large, such as 111M node number on ogbn-paper100M dataset \cite{OGB_dataset}, easily leading to out-of-memory on GPU or the long calculation time on CPU.

To demonstrate this challenge in deep graph clustering methods, we conduct experiments and theoretical analyses including comparison experiments, running cost evaluation, and complexity analyses. As shown in Table \ref{compare_table}. Concretely, fifteen important baselines are tested on four large graph datasets, including ogbn-arXiv, ogbn-products, Reddit, and ogbn-papers100M. The detailed statistics of datasets are summarized in Table \ref{statistics_information}. The baselines contain $k$-Means \cite{K-means}, DEC \cite{DEC}, DCN \cite{DCN}, node2vec \cite{node2vec}, DGI \cite{DGI}, AGE \cite{AGE}, MinCutPool \cite{mincutpool}, MVGRL \cite{MVGRL}, BGRL \cite{BYOL_graph}, GRACE \cite{GRACE}, ProGCL \cite{ProGCL}, AGC-DRR \cite{AGC-DRR}, DCRN \cite{DCRN}, S$^3$GC \cite{S3GC}, and Dink-Net \cite{dink_net}. From these experimental results, three conclusions are obtained as follows. Firstly, the classical method $k$-Means can scale to large graphs with million of nodes but can not achieve promising performance since the representation learning capability is limited. Secondly, most of the deep graph clustering methods like MVGRL \cite{MVGRL}, AGE \cite{AGE}, GRACE \cite{GRACE}, and DCRN \cite{DCRN} fail to run on large graphs. The reasons are that they either process the graph diffusion matrix or do not adopt batch training techniques. Thirdly, the scalable deep graph clustering methods S$^3$GC \cite{S3GC} and Dink-Net \cite{dink_net} can run on large graphs like ogbn-papers100M \cite{OGB_dataset} with 111 million nodes and achieve promising performance. S$^3$GC \cite{S3GC} conducts contrastive learning on the graph and performs $k$-Means on the learned node embeddings. In addition, Dink-Net \cite{dink_net} unifies the representation learning and clustering optimization via cluster dilation and cluster shrink loss functions. Besides, in Table \ref{table:complexity}, we analyze the time complexity and space complexity, and test the GPU memory costs and time costs of six deep graph clustering methods, including DEC \cite{DEC}, node2vec \cite{node2vec}, DGI \cite{DGI}, MVGRL \cite{MVGRL}, BGRL \cite{BYOL_graph}, S$^3$GC \cite{S3GC}, and Dink-Net \cite{dink_net}. 

Next, we discuss the potential techniques to improve the scalability of deep graph clustering methods. For the first part, i.e., encoder training on the large-scale graph, the possible techniques include the better sub-graph extraction method, more efficient batch training strategy, etc. For the second part, i.e., node clustering on the large-scale graph, the potential solutions contain the mini-batch $k$-Means \cite{mini-batch_kmeans}, finding cluster centers via the neural network \cite{dink_net}, calculating the clustering assignments in a mini-batch manner \cite{dink_net}, and bi-part graph clustering \cite{bipart_graph_1,bipart_graph_2}, etc. 

\begin{table*}[!t]
\centering
\caption{Discriminative capability testing experiments. Clustering performance of thirteen methods on six datasets}
\setlength{\tabcolsep}{2pt}
\small
\resizebox{0.9\linewidth}{!}{
% \scalebox{0.85}{
\begin{tabular}{ccccccccccccccc}
\hline
{\color[HTML]{000000} }                                        & {\color[HTML]{000000} }                                  & {\color[HTML]{000000} }                               & {\color[HTML]{000000} }                               & {\color[HTML]{000000} }                                 & {\color[HTML]{000000} }                                & {\color[HTML]{000000} }                                & {\color[HTML]{000000} }                                & {\color[HTML]{000000} }                                 & {\color[HTML]{000000} }                                & {\color[HTML]{000000} }                                 & {\color[HTML]{000000} }                                & {\color[HTML]{000000} }                                   & {\color[HTML]{000000} }                                & {\color[HTML]{000000} }                                 \\
\multirow{-2}{*}{{\color[HTML]{000000} \textbf{Dataset}}}      & \multirow{-2}{*}{{\color[HTML]{000000} \textbf{Metric}}} & \multirow{-2}{*}{{\color[HTML]{000000} \textbf{DEC}}} & \multirow{-2}{*}{{\color[HTML]{000000} \textbf{GAE}}} & \multirow{-2}{*}{{\color[HTML]{000000} \textbf{DAEGC}}} & \multirow{-2}{*}{{\color[HTML]{000000} \textbf{ARGA}}} & \multirow{-2}{*}{{\color[HTML]{000000} \textbf{SDCN}}} & \multirow{-2}{*}{{\color[HTML]{000000} \textbf{DFCN}}} & \multirow{-2}{*}{{\color[HTML]{000000} \textbf{MVGRL}}} & \multirow{-2}{*}{{\color[HTML]{000000} \textbf{MCGC}}} & \multirow{-2}{*}{{\color[HTML]{000000} \textbf{SCAGC}}} & \multirow{-2}{*}{{\color[HTML]{000000} \textbf{SCGC}}} & \multirow{-2}{*}{{\color[HTML]{000000} \textbf{AGC-DDR}}} & \multirow{-2}{*}{{\color[HTML]{000000} \textbf{DCRN}}} & \multirow{-2}{*}{{\color[HTML]{000000} \textbf{IDCRN}}} \\ \hline
{\color[HTML]{000000} }                                        & {\color[HTML]{000000} ACC}                               & {\color[HTML]{000000} 58.16}                          & {\color[HTML]{000000} 61.21}                          & {\color[HTML]{000000} 62.05}                            & {\color[HTML]{000000} 64.83}                           & {\color[HTML]{000000} 68.05}                           & {\color[HTML]{000000} 76.00}                           & {\color[HTML]{000000} 42.73}                            & {\color[HTML]{000000} 58.92}                           & {\color[HTML]{000000} 79.42}                            & {\color[HTML]{000000} 67.25}                           & {\color[HTML]{000000} 80.41}                              & {\color[HTML]{000000} 79.66}                           & {\color[HTML]{000000} 82.08}                            \\
{\color[HTML]{000000} }                                        & {\color[HTML]{000000} NMI}                               & {\color[HTML]{000000} 29.51}                          & {\color[HTML]{000000} 30.80}                          & {\color[HTML]{000000} 32.49}                            & {\color[HTML]{000000} 29.42}                           & {\color[HTML]{000000} 39.50}                           & {\color[HTML]{000000} 43.70}                           & {\color[HTML]{000000} 15.41}                            & {\color[HTML]{000000} 33.69}                           & {\color[HTML]{000000} 49.05}                            & {\color[HTML]{000000} 38.64}                           & {\color[HTML]{000000} 49.77}                              & {\color[HTML]{000000} 48.95}                           & {\color[HTML]{000000} 52.70}                            \\
{\color[HTML]{000000} }                                        & {\color[HTML]{000000} ARI}                               & {\color[HTML]{000000} 23.92}                          & {\color[HTML]{000000} 22.02}                          & {\color[HTML]{000000} 21.03}                            & {\color[HTML]{000000} 27.99}                           & {\color[HTML]{000000} 39.15}                           & {\color[HTML]{000000} 47.00}                           & {\color[HTML]{000000} 8.22}                             & {\color[HTML]{000000} 25.97}                           & {\color[HTML]{000000} 54.04}                            & {\color[HTML]{000000} 35.95}                           & {\color[HTML]{000000} 55.39}                              & {\color[HTML]{000000} 53.60}                           & {\color[HTML]{000000} 58.81}                            \\
\multirow{-4}{*}{{\color[HTML]{000000} \textbf{DBLP}}}         & {\color[HTML]{000000} F1}                                & {\color[HTML]{000000} 59.38}                          & {\color[HTML]{000000} 61.41}                          & {\color[HTML]{000000} 61.75}                            & {\color[HTML]{000000} 64.97}                           & {\color[HTML]{000000} 67.71}                           & {\color[HTML]{000000} 75.70}                           & {\color[HTML]{000000} 40.52}                            & {\color[HTML]{000000} 50.39}                           & {\color[HTML]{000000} 78.88}                            & {\color[HTML]{000000} 67.06}                           & {\color[HTML]{000000} 79.90}                              & {\color[HTML]{000000} 79.28}                           & {\color[HTML]{000000} 81.47}                            \\ \hline
{\color[HTML]{000000} }                                        & {\color[HTML]{000000} ACC}                               & {\color[HTML]{000000} 55.89}                          & {\color[HTML]{000000} 61.35}                          & {\color[HTML]{000000} 64.54}                            & {\color[HTML]{000000} 61.07}                           & {\color[HTML]{000000} 65.96}                           & {\color[HTML]{000000} 69.50}                           & {\color[HTML]{000000} 68.66}                            & {\color[HTML]{000000} 64.76}                           & {\color[HTML]{000000} 61.16}                            & {\color[HTML]{000000} 71.02}                           & {\color[HTML]{000000} 68.32}                              & {\color[HTML]{000000} 70.86}                           & {\color[HTML]{000000} 71.40}                             \\
{\color[HTML]{000000} }                                        & {\color[HTML]{000000} NMI}                               & {\color[HTML]{000000} 28.34}                          & {\color[HTML]{000000} 34.63}                          & {\color[HTML]{000000} 36.41}                            & {\color[HTML]{000000} 34.40}                           & {\color[HTML]{000000} 38.71}                           & {\color[HTML]{000000} 43.90}                           & {\color[HTML]{000000} 43.66}                            & {\color[HTML]{000000} 39.11}                           & {\color[HTML]{000000} 32.83}                            & {\color[HTML]{000000} 45.25}                           & {\color[HTML]{000000} 43.28}                              & {\color[HTML]{000000} 45.86}                           & {\color[HTML]{000000} 46.77}                            \\
{\color[HTML]{000000} }                                        & {\color[HTML]{000000} ARI}                               & {\color[HTML]{000000} 28.12}                          & {\color[HTML]{000000} 33.55}                          & {\color[HTML]{000000} 37.78}                            & {\color[HTML]{000000} 34.32}                           & {\color[HTML]{000000} 40.17}                           & {\color[HTML]{000000} 45.50}                           & {\color[HTML]{000000} 44.27}                            & {\color[HTML]{000000} 37.54}                           & {\color[HTML]{000000} 31.17}                            & {\color[HTML]{000000} 46.29}                           & {\color[HTML]{000000} 45.34}                              & {\color[HTML]{000000} 47.64}                           & {\color[HTML]{000000} 48.67}                            \\
\multirow{-4}{*}{{\color[HTML]{000000} \textbf{CiteSeer}}}     & {\color[HTML]{000000} F1}                                & {\color[HTML]{000000} 52.62}                          & {\color[HTML]{000000} 57.36}                          & {\color[HTML]{000000} 62.20}                            & {\color[HTML]{000000} 58.23}                           & {\color[HTML]{000000} 63.62}                           & {\color[HTML]{000000} 64.30}                           & {\color[HTML]{000000} 63.71}                            & {\color[HTML]{000000} 59.64}                           & {\color[HTML]{000000} 56.82}                            & {\color[HTML]{000000} 62.24}                           & {\color[HTML]{000000} 64.82}                              & {\color[HTML]{000000} 65.83}                           & {\color[HTML]{000000} 66.27}                            \\ \hline
{\color[HTML]{000000} }                                        & {\color[HTML]{000000} ACC}                               & {\color[HTML]{000000} 84.33}                          & {\color[HTML]{000000} 84.52}                          & {\color[HTML]{000000} 86.94}                            & {\color[HTML]{000000} 86.29}                           & {\color[HTML]{000000} 90.45}                           & {\color[HTML]{000000} 90.90}                           & {\color[HTML]{000000} 86.73}                            & {\color[HTML]{000000} 91.64}                           & {\color[HTML]{000000} 91.83}                            & {\color[HTML]{000000} 89.80}                           & {\color[HTML]{000000} 92.55}                              & {\color[HTML]{000000} 91.93}                           & {\color[HTML]{000000} 92.58}                            \\
{\color[HTML]{000000} }                                        & {\color[HTML]{000000} NMI}                               & {\color[HTML]{000000} 54.54}                          & {\color[HTML]{000000} 55.38}                          & {\color[HTML]{000000} 56.18}                            & {\color[HTML]{000000} 56.21}                           & {\color[HTML]{000000} 68.31}                           & {\color[HTML]{000000} 69.40}                           & {\color[HTML]{000000} 60.87}                            & {\color[HTML]{000000} 70.71}                           & {\color[HTML]{000000} 71.28}                            & {\color[HTML]{000000} 66.60}                           & {\color[HTML]{000000} 72.89}                              & {\color[HTML]{000000} 71.56}                           & {\color[HTML]{000000} 73.17}                            \\
{\color[HTML]{000000} }                                        & {\color[HTML]{000000} ARI}                               & {\color[HTML]{000000} 60.64}                          & {\color[HTML]{000000} 59.46}                          & {\color[HTML]{000000} 59.35}                            & {\color[HTML]{000000} 63.37}                           & {\color[HTML]{000000} 73.91}                           & {\color[HTML]{000000} 74.90}                           & {\color[HTML]{000000} 65.07}                            & {\color[HTML]{000000} 76.63}                           & {\color[HTML]{000000} 77.29}                            & {\color[HTML]{000000} 72.37}                           & {\color[HTML]{000000} 79.08}                              & {\color[HTML]{000000} 77.56}                           & {\color[HTML]{000000} 79.18}                            \\
\multirow{-4}{*}{{\color[HTML]{000000} \textbf{ACM}}}          & {\color[HTML]{000000} F1}                                & {\color[HTML]{000000} 84.51}                          & {\color[HTML]{000000} 84.65}                          & {\color[HTML]{000000} 87.07}                            & {\color[HTML]{000000} 86.31}                           & {\color[HTML]{000000} 90.42}                           & {\color[HTML]{000000} 90.80}                           & {\color[HTML]{000000} 86.85}                            & {\color[HTML]{000000} 91.70}                           & {\color[HTML]{000000} 91.84}                            & {\color[HTML]{000000} 89.74}                           & {\color[HTML]{000000} 92.55}                              & {\color[HTML]{000000} 91.94}                           & {\color[HTML]{000000} 92.60}                            \\ \hline
{\color[HTML]{000000} }                                        & {\color[HTML]{000000} ACC}                               & {\color[HTML]{000000} 47.22}                          & {\color[HTML]{000000} 71.57}                          & {\color[HTML]{000000} 76.44}                            & {\color[HTML]{000000} 69.28}                           & {\color[HTML]{000000} 53.44}                           & {\color[HTML]{000000} 76.88}                           & {\color[HTML]{000000} 45.19}                            & {\color[HTML]{000000} OOM}                             & {\color[HTML]{000000} 75.25}                            & {\color[HTML]{000000} 77.48}                           & {\color[HTML]{000000} 78.11}                              & {\color[HTML]{000000} 79.94}                           & {\color[HTML]{000000} 80.17}                            \\
{\color[HTML]{000000} }                                        & {\color[HTML]{000000} NMI}                               & {\color[HTML]{000000} 37.35}                          & {\color[HTML]{000000} 62.13}                          & {\color[HTML]{000000} 65.57}                            & {\color[HTML]{000000} 58.36}                           & {\color[HTML]{000000} 44.85}                           & {\color[HTML]{000000} 69.21}                           & {\color[HTML]{000000} 36.89}                            & {\color[HTML]{000000} OOM}                             & {\color[HTML]{000000} 67.18}                            & {\color[HTML]{000000} 67.67}                           & {\color[HTML]{000000} 72.21}                              & {\color[HTML]{000000} 73.70}                           & {\color[HTML]{000000} 74.32}                            \\
{\color[HTML]{000000} }                                        & {\color[HTML]{000000} ARI}                               & {\color[HTML]{000000} 18.59}                          & {\color[HTML]{000000} 48.82}                          & {\color[HTML]{000000} 59.39}                            & {\color[HTML]{000000} 44.18}                           & {\color[HTML]{000000} 31.21}                           & {\color[HTML]{000000} 58.98}                           & {\color[HTML]{000000} 18.79}                            & {\color[HTML]{000000} OOM}                             & {\color[HTML]{000000} 56.86}                            & {\color[HTML]{000000} 58.48}                           & {\color[HTML]{000000} 61.15}                              & {\color[HTML]{000000} 63.69}                           & {\color[HTML]{000000} 64.10}                            \\
\multirow{-4}{*}{{\color[HTML]{000000} \textbf{Amazon-Photo}}} & {\color[HTML]{000000} F1}                                & {\color[HTML]{000000} 46.71}                          & {\color[HTML]{000000} 68.08}                          & {\color[HTML]{000000} 69.97}                            & {\color[HTML]{000000} 64.30}                           & {\color[HTML]{000000} 50.66}                           & {\color[HTML]{000000} 71.58}                           & {\color[HTML]{000000} 39.65}                            & {\color[HTML]{000000} OOM}                             & {\color[HTML]{000000} 72.77}                            & {\color[HTML]{000000} 72.22}                           & {\color[HTML]{000000} 72.72}                              & {\color[HTML]{000000} 73.82}                           & {\color[HTML]{000000} 74.01}                            \\ \hline
{\color[HTML]{000000} }                                        & {\color[HTML]{000000} ACC}                               & {\color[HTML]{000000} 60.14}                          & {\color[HTML]{000000} 62.09}                          & {\color[HTML]{000000} 68.73}                            & {\color[HTML]{000000} 65.26}                           & {\color[HTML]{000000} 64.20}                           & {\color[HTML]{000000} 68.89}                           & {\color[HTML]{000000} 67.01}                            & {\color[HTML]{000000} 60.97}                           & {\color[HTML]{000000} OOM}                              & {\color[HTML]{000000} 63.14}                           & {\color[HTML]{000000} -}                                  & {\color[HTML]{000000} 69.87}                           & {\color[HTML]{000000} 70.02}                            \\
{\color[HTML]{000000} }                                        & {\color[HTML]{000000} NMI}                               & {\color[HTML]{000000} 22.44}                          & {\color[HTML]{000000} 23.84}                          & {\color[HTML]{000000} 28.26}                            & {\color[HTML]{000000} 24.80}                           & {\color[HTML]{000000} 22.87}                           & {\color[HTML]{000000} 31.43}                           & {\color[HTML]{000000} 31.59}                            & {\color[HTML]{000000} 33.39}                           & {\color[HTML]{000000} OOM}                              & {\color[HTML]{000000} 27.77}                           & {\color[HTML]{000000} -}                                  & {\color[HTML]{000000} 32.20}                           & {\color[HTML]{000000} 33.29}                            \\
{\color[HTML]{000000} }                                        & {\color[HTML]{000000} ARI}                               & {\color[HTML]{000000} 19.55}                          & {\color[HTML]{000000} 20.62}                          & {\color[HTML]{000000} 29.84}                            & {\color[HTML]{000000} 24.35}                           & {\color[HTML]{000000} 22.30}                           & {\color[HTML]{000000} 30.64}                           & {\color[HTML]{000000} 29.42}                            & {\color[HTML]{000000} 29.25}                           & {\color[HTML]{000000} OOM}                              & {\color[HTML]{000000} 24.73}                           & {\color[HTML]{000000} -}                                  & {\color[HTML]{000000} 31.41}                           & {\color[HTML]{000000} 32.67}                            \\
\multirow{-4}{*}{{\color[HTML]{000000} \textbf{PubMed}}}       & {\color[HTML]{000000} F1}                                & {\color[HTML]{000000} 61.49}                          & {\color[HTML]{000000} 61.37}                          & {\color[HTML]{000000} 68.23}                            & {\color[HTML]{000000} 65.69}                           & {\color[HTML]{000000} 65.01}                           & {\color[HTML]{000000} 68.10}                           & {\color[HTML]{000000} 67.07}                            & {\color[HTML]{000000} 59.84}                           & {\color[HTML]{000000} OOM}                              & {\color[HTML]{000000} 62.27}                           & {\color[HTML]{000000} -}                                  & {\color[HTML]{000000} 68.94}                           & {\color[HTML]{000000} 69.19}                            \\ \hline
{\color[HTML]{000000} }                                        & {\color[HTML]{000000} ACC}                               & {\color[HTML]{000000} 31.92}                          & {\color[HTML]{000000} 29.60}                          & {\color[HTML]{000000} 34.35}                            & {\color[HTML]{000000} 22.07}                           & {\color[HTML]{000000} 26.67}                           & {\color[HTML]{000000} 37.51}                           & {\color[HTML]{000000} 31.52}                            & {\color[HTML]{000000} 29.08}                           & {\color[HTML]{000000} OOM}                              & {\color[HTML]{000000} OOM}                             & {\color[HTML]{000000} -}                                  & {\color[HTML]{000000} 38.80}                           & {\color[HTML]{000000} 39.45}                            \\
{\color[HTML]{000000} }                                        & {\color[HTML]{000000} NMI}                               & {\color[HTML]{000000} 41.67}                          & {\color[HTML]{000000} 45.82}                          & {\color[HTML]{000000} 49.16}                            & {\color[HTML]{000000} 41.28}                           & {\color[HTML]{000000} 37.38}                           & {\color[HTML]{000000} 51.30}                           & {\color[HTML]{000000} 48.99}                            & {\color[HTML]{000000} 36.86}                           & {\color[HTML]{000000} OOM}                              & {\color[HTML]{000000} OOM}                             & {\color[HTML]{000000} -}                                  & {\color[HTML]{000000} 51.91}                           & {\color[HTML]{000000} 52.83}                            \\
{\color[HTML]{000000} }                                        & {\color[HTML]{000000} ARI}                               & {\color[HTML]{000000} 16.98}                          & {\color[HTML]{000000} 17.84}                          & {\color[HTML]{000000} 22.60}                            & {\color[HTML]{000000} 12.38}                           & {\color[HTML]{000000} 13.63}                           & {\color[HTML]{000000} 24.46}                           & {\color[HTML]{000000} 19.11}                            & {\color[HTML]{000000} 13.15}                           & {\color[HTML]{000000} OOM}                              & {\color[HTML]{000000} OOM}                             & {\color[HTML]{000000} -}                                  & {\color[HTML]{000000} 25.25}                           & {\color[HTML]{000000} 25.97}                            \\
\multirow{-4}{*}{{\color[HTML]{000000} \textbf{CoraFULL}}}     & {\color[HTML]{000000} F1}                                & {\color[HTML]{000000} 27.71}                          & {\color[HTML]{000000} 25.95}                          & {\color[HTML]{000000} 26.96}                            & {\color[HTML]{000000} 18.85}                           & {\color[HTML]{000000} 22.14}                           & {\color[HTML]{000000} 31.22}                           & {\color[HTML]{000000} 26.51}                            & {\color[HTML]{000000} 22.90}                           & {\color[HTML]{000000} OOM}                              & {\color[HTML]{000000} OOM}                             & {\color[HTML]{000000} -}                                  & {\color[HTML]{000000} 31.68}                           & {\color[HTML]{000000} 32.58}                            \\ \hline
\end{tabular}}
\label{discri_table}
\end{table*}

% VIS
\begin{figure*}[!t]
\centering
% \tiny
\begin{minipage}{0.15\linewidth}
\centerline{\includegraphics[width=\textwidth]{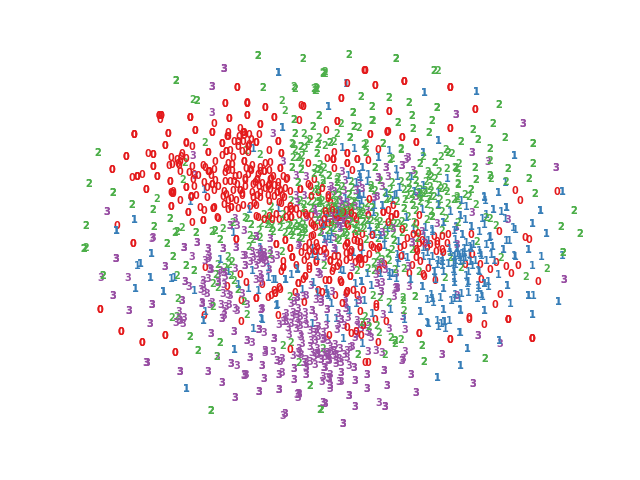}}
\vspace{5pt}
\centerline{\includegraphics[width=\textwidth]{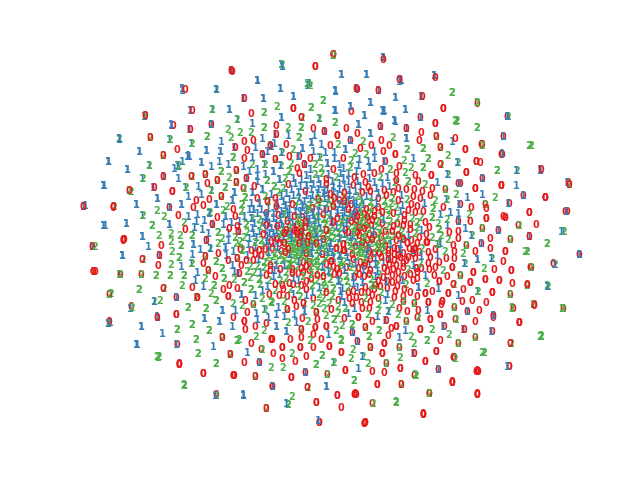}}
\vspace{5pt}
\centerline{(a) Raw Data}
\end{minipage}
\begin{minipage}{0.15\linewidth}
\centerline{\includegraphics[width=\textwidth]{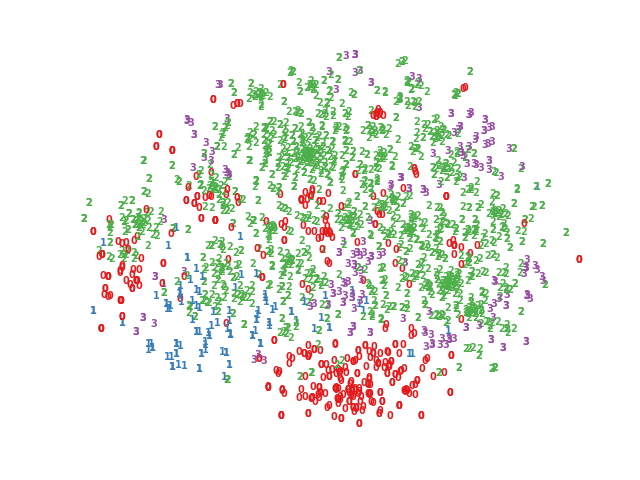}}
\vspace{5pt}
\centerline{\includegraphics[width=\textwidth]{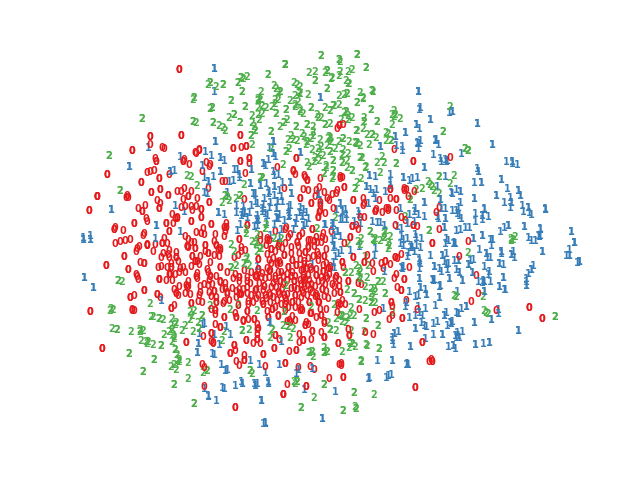}}
\vspace{5pt}
\centerline{(b) DEC}
\end{minipage}
\begin{minipage}{0.15\linewidth}
\centerline{\includegraphics[width=\textwidth]{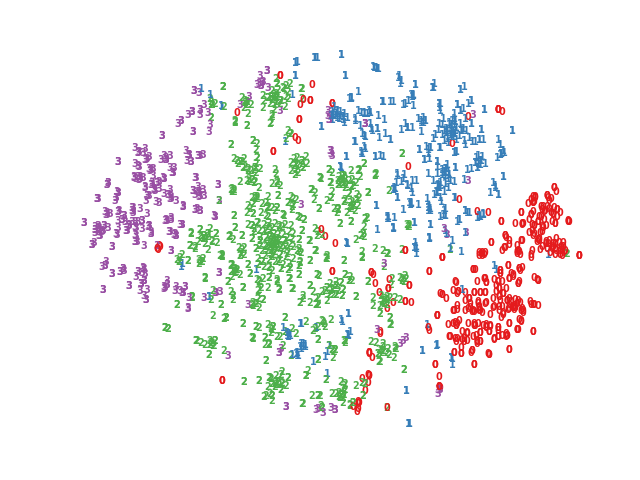}}
\vspace{5pt}
\centerline{\includegraphics[width=\textwidth]{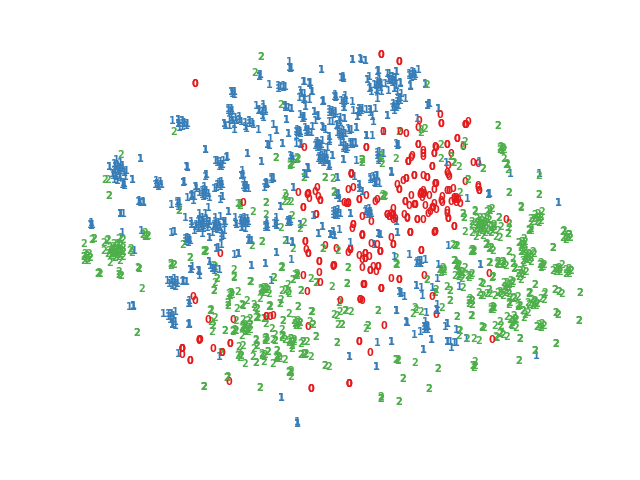}}
\vspace{5pt}
\centerline{(c) GAE}
\end{minipage}
\begin{minipage}{0.15\linewidth}
\centerline{\includegraphics[width=\textwidth]{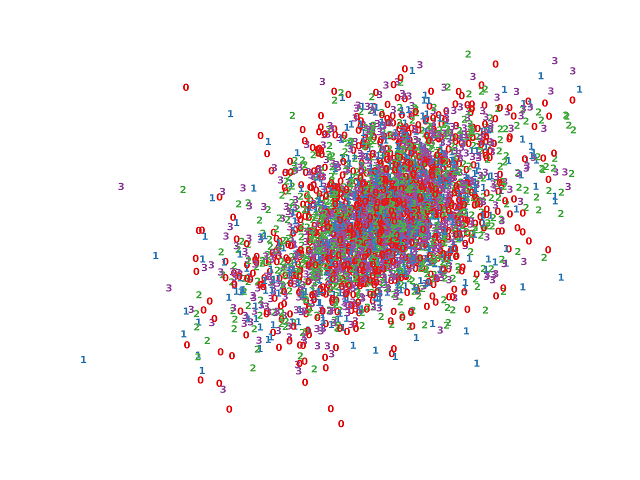}}
\vspace{5pt}
\centerline{\includegraphics[width=\textwidth]{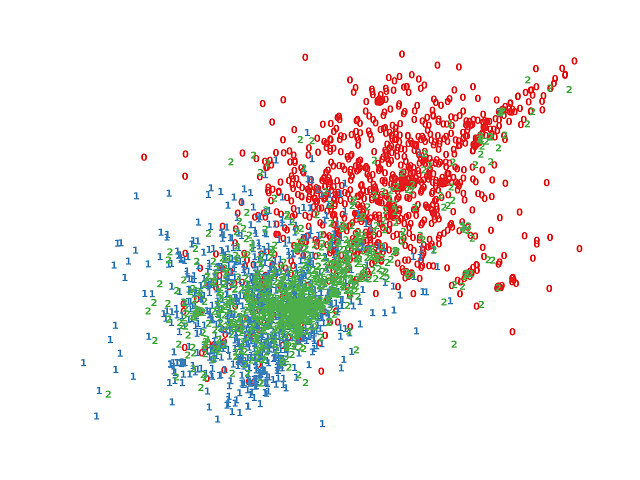}}
\vspace{5pt}
\centerline{(d) ARGA}
\end{minipage}
\begin{minipage}{0.15\linewidth}
\centerline{\includegraphics[width=\textwidth]{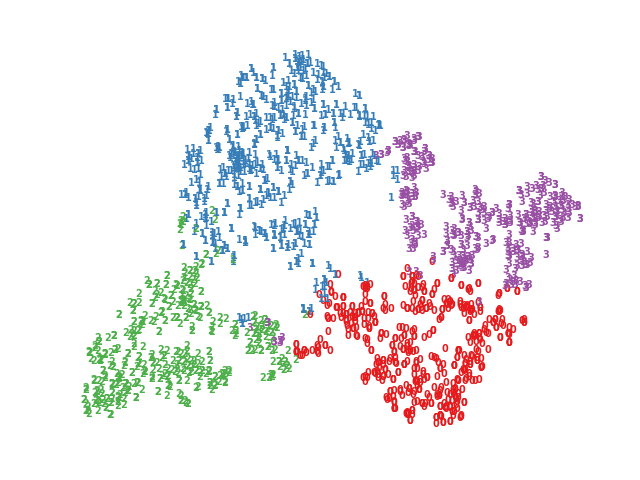}}
\vspace{5pt}
\centerline{\includegraphics[width=\textwidth]{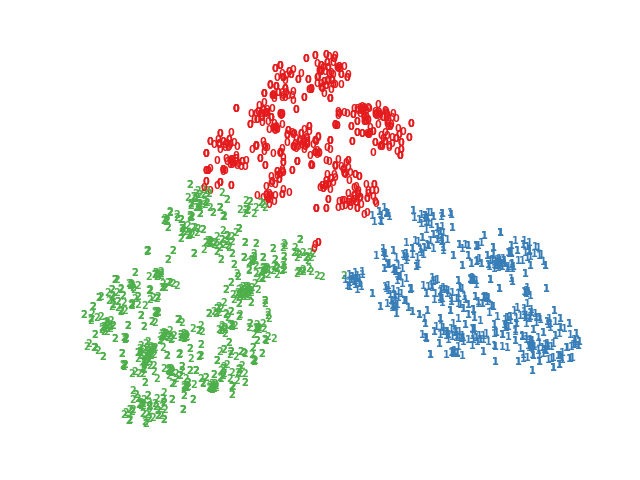}}
\vspace{5pt}
\centerline{(e) DFCN}
\end{minipage}
\begin{minipage}{0.15\linewidth}
\centerline{\includegraphics[width=\textwidth]{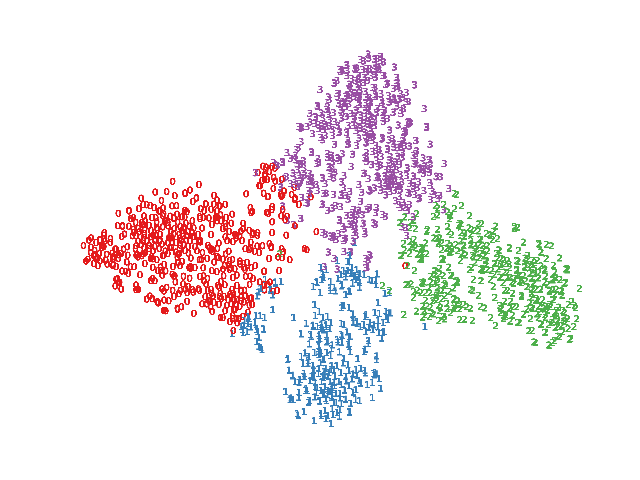}}
\vspace{5pt}
\centerline{\includegraphics[width=\textwidth]{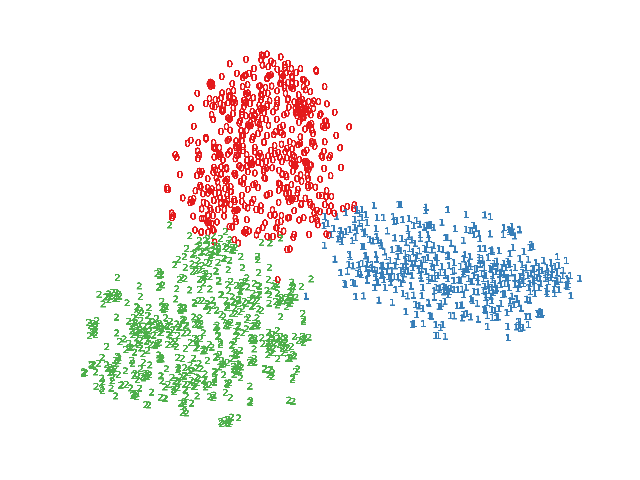}}
\vspace{5pt}
\centerline{(f) IDCRN}
\end{minipage}
\caption{Discriminative capability testing experiments. $t$-SNE visualization of node embeddings. The compared methods include DEC \cite{DEC}, GAE \cite{GAE}, ARGA \cite{ARGA}, DFCN \cite{DFCN}, and IDCRN \cite{IDCRN}. The first and second row is the result on ACM and DBLP dataset.}
\label{tsne_fig}
\end{figure*}

% MOTIVATION ON CITE AND DBLP
\begin{figure}[!t]
\centering
\small
\begin{minipage}{0.24\linewidth}
\centerline{\includegraphics[width=\textwidth]{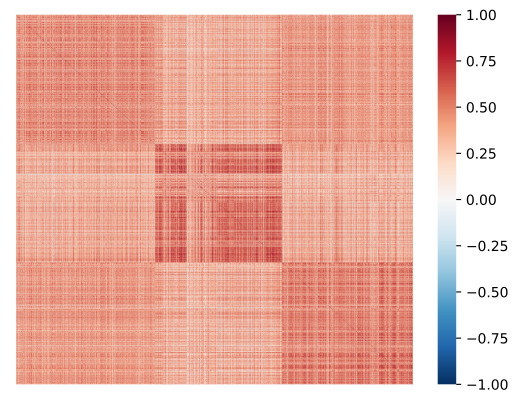}}
\vspace{5pt}
\centerline{\includegraphics[width=\textwidth]{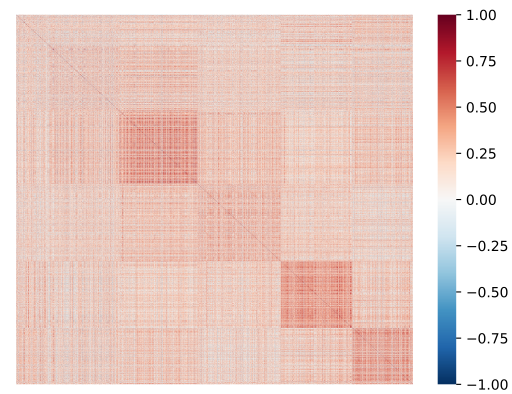}}
\vspace{5pt}
\centerline{(a) GAE}
\end{minipage}
\begin{minipage}{0.24\linewidth}
\centerline{\includegraphics[width=\textwidth]{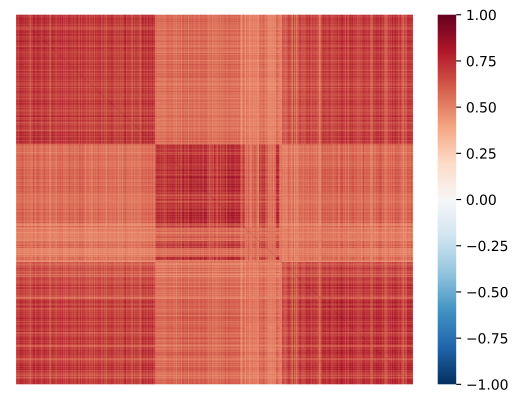}}
\vspace{5pt}
\centerline{\includegraphics[width=\textwidth]{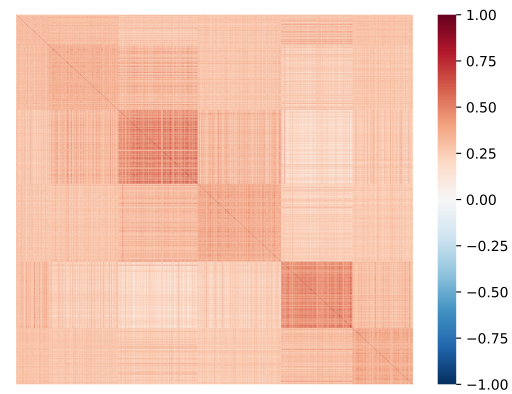}}
\vspace{5pt}
\centerline{(b) MVGRL}
\end{minipage}
\begin{minipage}{0.24\linewidth}
\centerline{\includegraphics[width=\textwidth]{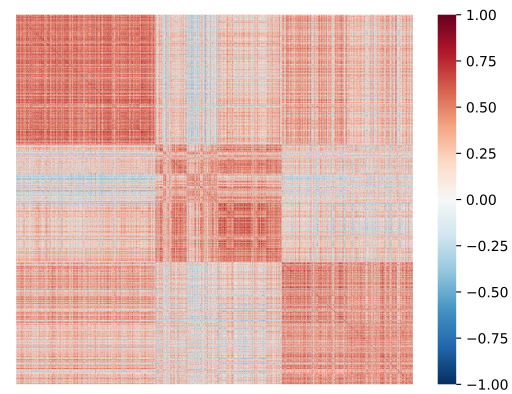}}
\vspace{5pt}
\centerline{\includegraphics[width=\textwidth]{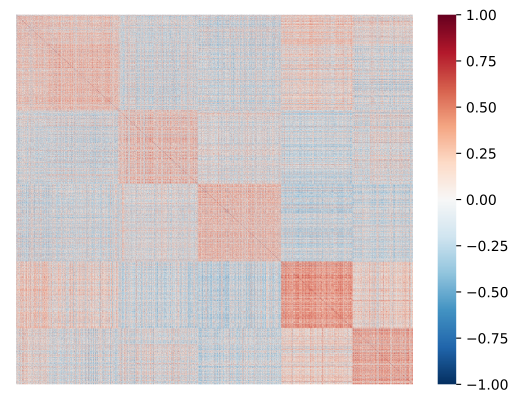}}
\vspace{5pt}
\centerline{(c)SDCN}
\end{minipage}
\begin{minipage}{0.24\linewidth}
\centerline{\includegraphics[width=\textwidth]{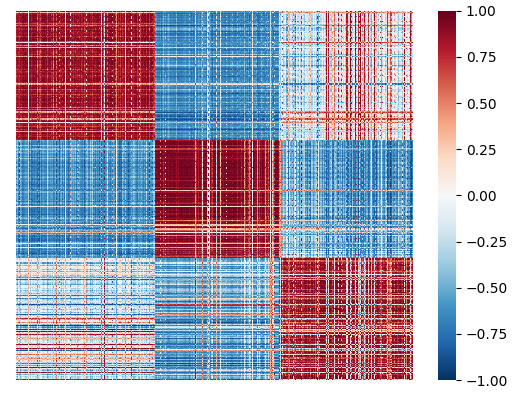}}
\vspace{5pt}
\centerline{\includegraphics[width=\textwidth]{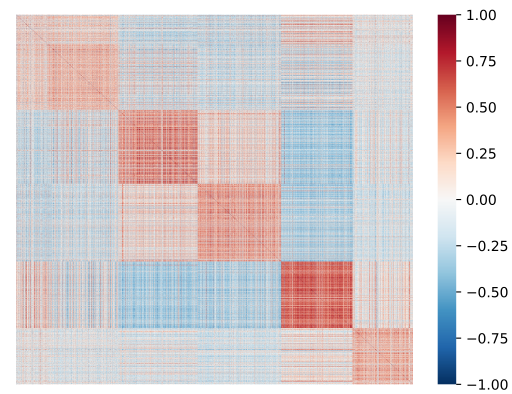}}
\vspace{5pt}
\centerline{(d) IDCRN}
\end{minipage}
\caption{Discriminative capability testing experiments. Visualization of pairwise sample similarity. The compared methods are GAE \cite{GAE}, MVGRL \cite{MVGRL}, SDCN \cite{SDCN}, and IDCRN \cite{IDCRN}. The first row and second row is the result on ACM and CiteSeer dataset.}
\label{sim_fig}  
\end{figure}

\subsection{Discriminative Capability}
One key factor of promising clustering performance is to learn discriminative node embeddings, which help the clustering methods better group nodes into different clusters. Therefore, the discriminative capability of the network is vital in self-supervised learning. \cite{DCRN} observe the representation collapse problem, i.e., the encoder tends to embed all samples into the same cluster in deep graph clustering methods. To demonstrate this problem, we conduct extensive experiments, including three types: performance comparison experiments, similarity visualization experiments, and embedding visualization experiments. 

At first, as shown in Table \ref{discri_table}, we conduct performance comparison experiments on six datasets, including DBLP, CiteSeer, ACM, Amazon-Photo, PubMed, and CoraFull datasets. The detailed information of these datasets is listed in Table \ref{statistics_information}. We compare thirteen state-of-the-art methods including DEC \cite{DEC}, GAE \cite{GAE}, DAEGC \cite{DAEGC}, ARGA \cite{ARGA}, SDCN \cite{SDCN}, DFCN \cite{DFCN}, MVGRL \cite{MVGRL}, MCGC \cite{MCGC}, SCAGC \cite{SCAGC}, SCGC \cite{SCGC}, AGC-DDR \cite{AGC-DRR}, DCRN \cite{DCRN}, and IDCRN \cite{IDCRN}. From these experimental results, we have three observations as follows. Firstly, the performance of deep clustering methods such as DEC \cite{DEC} is limited since it omits the graph structure of graph data. Secondly, various deep reconstructive graph clustering methods such as GAE \cite{GAE}, DAEGC \cite{DAEGC}, SDCN \cite{SDCN}, DFCN \cite{DFCN} significantly improve the clustering performance since they reconstruct the graph information and improve the discriminative capability of networks. Thirdly, the recent contrastive methods like MVGRL \cite{MVGRL}, SCGC \cite{SCGC}, and DCRN \cite{DCRN} further enhance the sample discriminative capability via pulling together positive sample pairs and pushing away the negative sample pairs, achieving the most promising performance.

% various contrastive deep graph clustering. 

% Thirdly, . 

% , DCRN \cite{DCRN} and IDCRN \cite{IDCRN} enhance these 

% From these experimental results, we find that DCRN 

In addition, to intuitively demonstrate this challenge, we conduct the $t$-SNE visualization on the learned node embeddings. As shown in Figure \ref{tsne_fig}, Three observations can be found as follows. Firstly, in the raw data, we can not observe any cluster shape in the latent space. Secondly, the deep clustering method DEC \cite{DEC}, reconstructive deep graph clustering \cite{GAE,DFCN}, and adversarial deep graph clustering method ARGA \cite{ARGA} can reveal part of the clustering distribution. Thirdly, the sample discriminative capability of the contrastive methods IDCRN \cite{IDCRN} is strong, revealing the clustering distribution well. We consider visualizing the learned node embeddings as one of the intuitive ways to evaluate the discriminative capability of deep graph clustering methods.

% Thirdly, 
% In addition, 

Furthermore, we calculate and visualize the pair-wise similarity of the learned node embeddings. The results are demonstrated in Figure \ref{sim_fig}. Here, we reorder the nodes and gather the nodes with the same labels on the same side. Besides, the red pixel denotes the high similarity, and the blue pixel denotes the low similarity. From the figure, two conclusions are drawn. Firstly, GAE \cite{GAE} and MVGRL \cite{MVGRL} have a trend to the representation collapse problem since all paired sample similarities are high. Secondly, SDCN \cite{SDCN} and IDCRN \cite{IDCRN} alleviate the representation collapse problem via the delivery operator and dual correlation reduction. As shown in Figure \ref{sim_fig} (c) and Figure \ref{sim_fig} (d), the similarities between the samples in the same cluster are relatively high. In contrast, the similarities between the samples in different clusters are relatively low. We think this similarity visualization method can be an excellent tool to test the sample discriminative capability of the deep graph clustering method.

Recently, we have witnessed the fast development of self-supervised learning on graphs \cite{ssl_graph}. The reconstructive deep graph clustering methods \cite{GAE,DAEGC,DFCN,SDCN,O2MAC} reconstruct the graph information such as node attribute and graph structure. Besides, the adversarial methods \cite{ARGA,AGAE,AGC-DRR} generate the adversarial samples and discriminate them from the real samples. In addition, the contrastive methods \cite{DCRN,AGC-DRR,SCAGC,dink_net} pull together the positive sample pairs while pushing away the negative sample pairs. Previous researchers designed different promising pre-text tasks to enhance the sample discriminative capability of networks. In the future, to further enhance the discriminative capability, the hard sample mining strategies \cite{ProGCL,HSAN} will be an exciting future direction. 

% Recently, more and more methods \cite{DCRN,AGC-DRR,SCAGC} utilize the contrastive learning paradigm to pull together the positive samples while pushing away the negative ones, thus improving the sample discriminative capability of the network.

\subsection{Unknown Cluster Number}
Most of the existing deep graph clustering methods consider the cluster number as a given value, which is usually the same as the classes of the ground truth. Benefiting from this setting, recent deep graph clustering methods can achieve promising performance. The remarkable success of recent deep graph clustering algorithms relies on the pre-defined cluster number. However, in most real-world scenarios, the cluster number is an unknown value. For example, in Internet social networks, the users can be grouped into several clusters, but we can not know the specific number of user groups. In addition, the number of user groups is the most important in the user grouping algorithm. Therefore, we think the clustering performance of the existing deep graph clustering methods is overestimated.

% in Figure \ref{cluster_number}.
\begin{figure}[h]
\centering
\small
\begin{minipage}{0.49\linewidth}
\centerline{\includegraphics[width=1\textwidth]{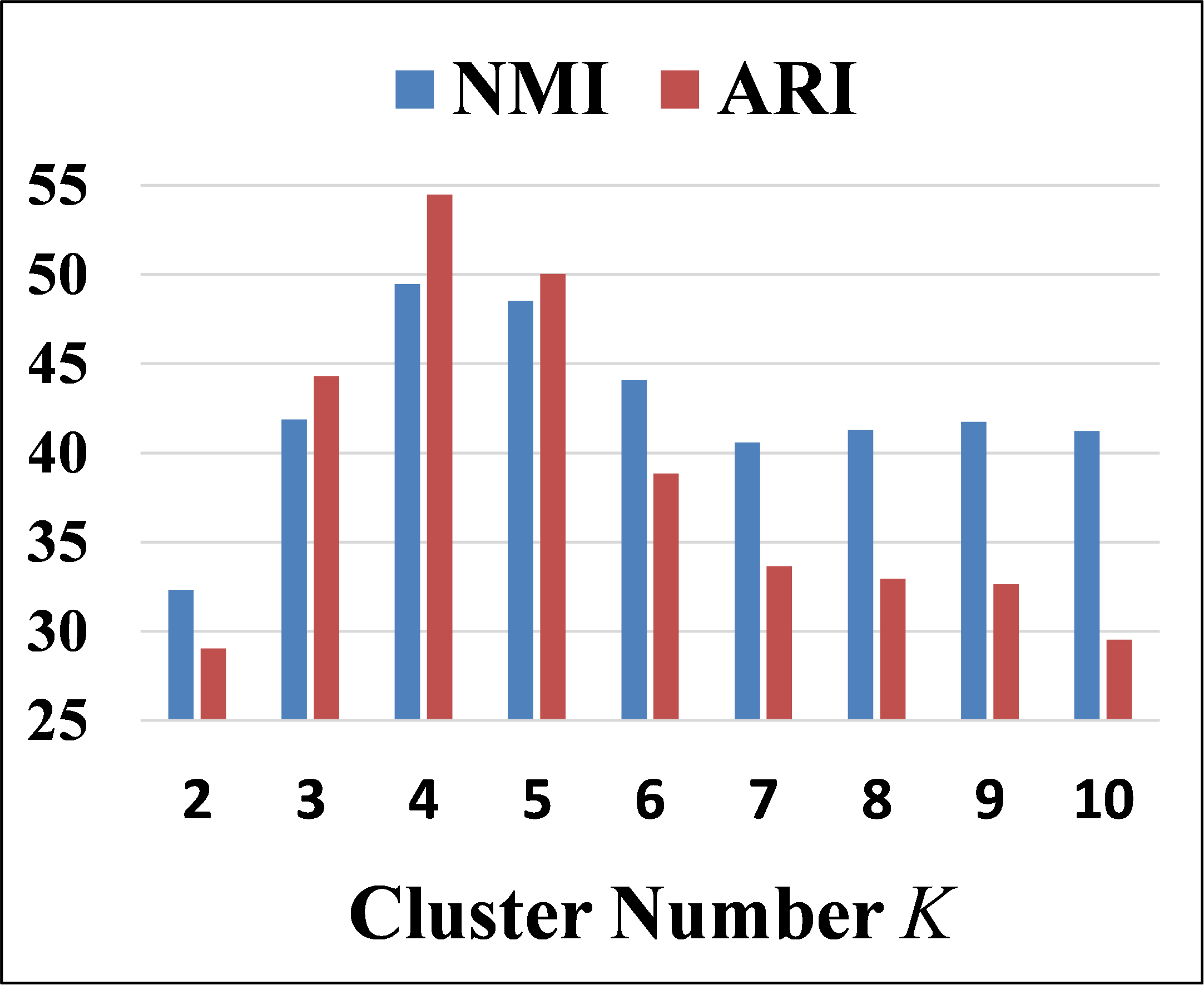}}
\centerline{(a) DBLP}
\vspace{3pt}
\end{minipage}
\begin{minipage}{0.49\linewidth}
\centerline{\includegraphics[width=1\textwidth]{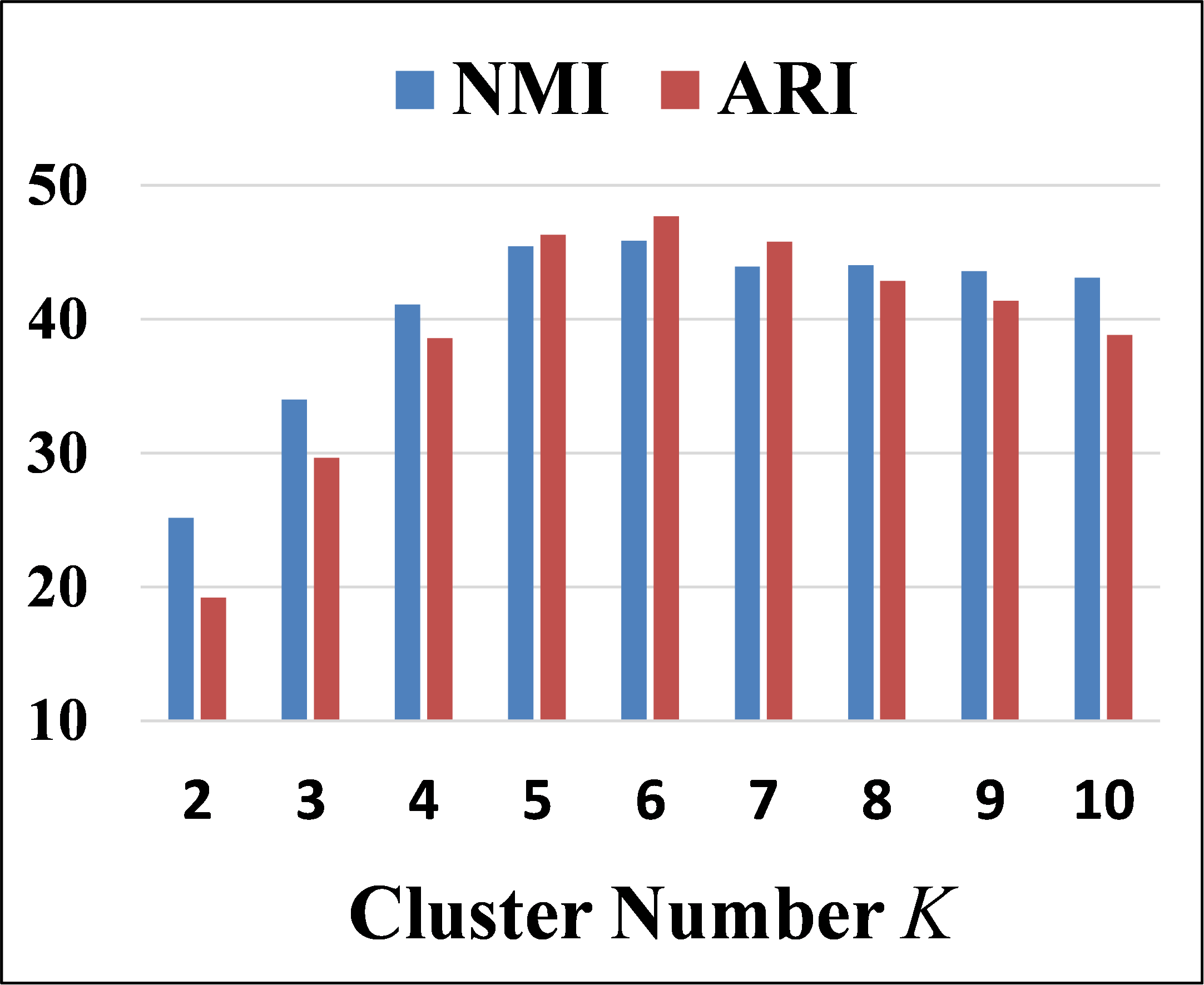}}
\centerline{(b) CiteSeer}
\vspace{3pt}
\end{minipage}
\caption{Unknown cluster number experiments. Clustering performance of DCRN \cite{DCRN} with different cluster numbers $K \in [2,10]$ on Cora and CiteSeer datasets.}
\label{cluster_number}
\end{figure}

\begin{figure*}[!t]
\centering
% \small
\begin{minipage}{0.9\linewidth}
\centerline{\includegraphics[width=1\textwidth]{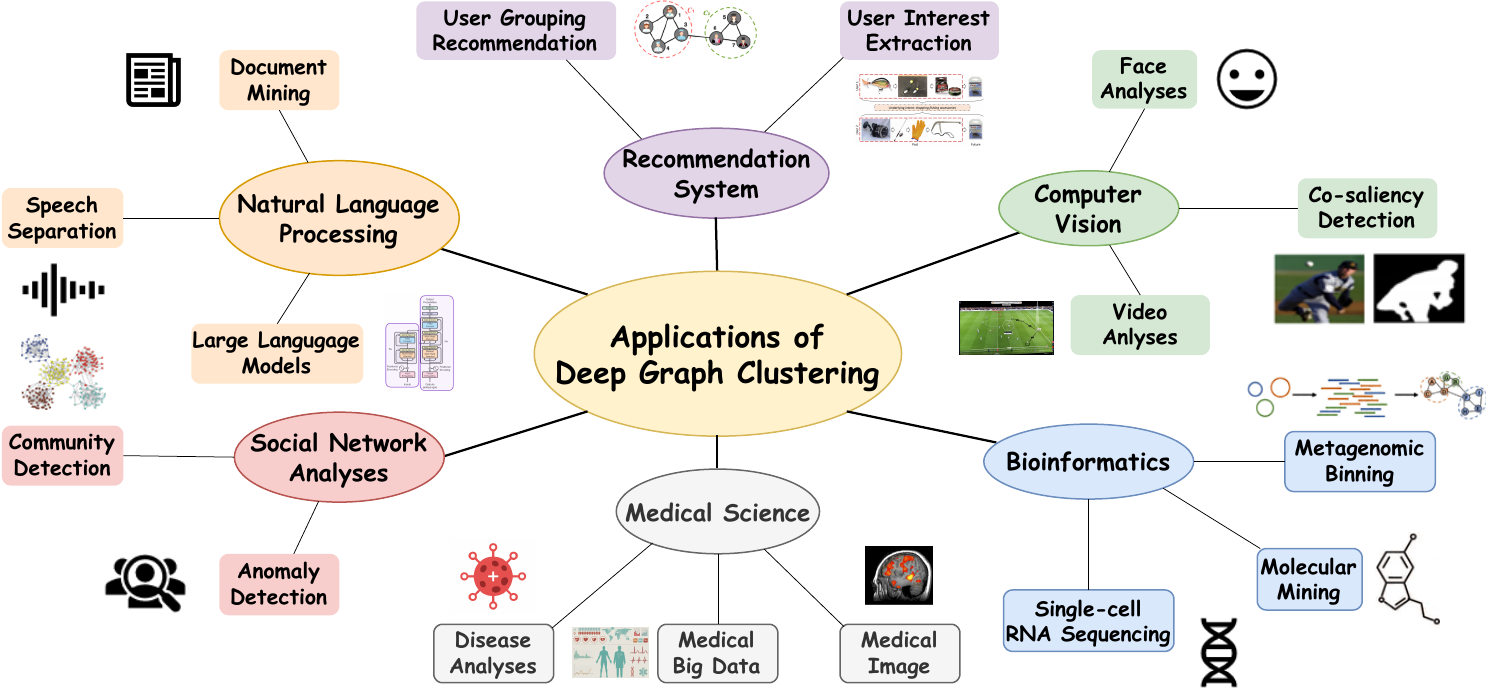}}
\end{minipage}
\caption{Applications of deep graph clustering in six domains: natural language processing, computer vision, social network analyses, recommendation systems, bioinformatics, and medical science. }
\label{application_figure}
\end{figure*}

In order to verify our suspect, we conduct experiments on two datasets, including DBLP and CiteSeer. Concretely, we adopt DCRN \cite{DCRN} as the baseline and evaluate its clustering performance with NMI and ARI metrics using different cluster numbers ranging from 2 to 10. The experimental results are demonstrated in Figure \ref{cluster_number}. We carefully analyze the results the draw two conclusions as follows. Firstly, the baseline DCRN achieves promising and best performance when the cluster number equals the class number of ground truth. Here, the sample class number of the DBLP dataset is 4, and the sample class number of CiteSeer is 6. Secondly, other cluster numbers will lead to performance degeneration. When the cluster number reduces, many clusters will collapse into one cluster. Besides, when the cluster number increases, the large cluster will be split into many small clusters. Therefore, the cluster number is crucial for the clustering performance, and designing the deep graph clustering methods without the pre-defined cluster number will be a challenging and meaningful direction.

% From the experimental results, we observe that the wrong cluster number will lead to performance degeneration. 

To solve this problem, one potential solution is to perform clustering based on density \cite{DPCA}. Concretely, the clustering network should be designed to assign the high-density data area as the cluster centers and regard the low-density data area as the decision borders. Besides, another potential solution technology is deep reinforcement learning \cite{RL_survey,liuyue_RGC}. To be specific, in an unsupervised scenario, recognizing the cluster number in the graph data can be modeled as the Markov decision process and be handled with deep reinforcement neural networks. More cluster number determination methods refer to \cite{determine_k}. In the next section, we will introduce the applications of deep graph clustering in four domains.

\section{Application \& Open Resource} \label{application_section}
\subsection{Application}
In recent years, we have witnessed the fast growth of deep graph clustering. Thanks to the researchers in this field, promising deep graph clustering methods are increasingly proposed. Benefiting from the strong data partitioning capability, deep graph clustering has been applied to various real-world application domains, such as natural language processing, computer vision, social network analyses, recommendation systems, bioinformatics, medical science, etc. The applications are demonstrated in Figure \ref{application_figure}. 

Next, we introduce the applications in detail. In the computer vision domain, deep graph clustering methods are applied to face analysis \cite{face_1,face_2}, co-saliency detection \cite{co-saliency_1} and video analyses \cite{video_cluster_1}. Besides, document mining \cite{document_1}, speech separation \cite{speech_separation_1}, and large language models \cite{cluster_LLM_1} are essential applications of deep graph clustering in the natural language processing area. In addition, deep graph clustering is crucial for social network analyses. It can be used to conduct community detection \cite{ComE,Gemsec,CNRL,community_detection_1} and anomaly detection \cite{AD_1,AD_2}. Similarly, deep graph clustering demonstrates the high application values in the recommendation systems. To be specific, it can help user grouping recommendations and user intent extraction \cite{ICLRec}. Apart from social data mining, deep graph clustering is also vital to bioinformatics and medical science. Concretely, the applications in the bioinformatics field include molecular mining \cite{molecular_1,molecular_2}, metagenomic binning \cite{Repbin}, single-cell RNA sequencing \cite{scTAG}, etc. Also, in the medical science domain, deep graph clustering methods are adopted in disease analyses \cite{disease_analyses_1}, medical big data \cite{medical_big_data_1}, and medical image \cite{medical_image_1}. In the future, we hope the researchers will further solve the challenges and apply deep graph clustering methods to a broader range and more significant fields.

% different domains of real-world applications,
% document mining \cite{document_1}, speech separation \cite{speech_separation_1}, 

% such as social network analysis, computer vision, natural language processing, bioinformatics, etc. 
% co-saliency detection \cite{co-saliency_1}

% As shown in Figure \ref{application_figure}, the specific applications of deep graph clustering contain face analysis \cite{face_1,face_2}, anomaly detection \cite{AD_1,AD_2}, 

% , community detection \cite{ComE,Gemsec,CNRL,community_detection_1}, 

\subsection{Open Resource}
In order to construct an open, active, and collaborative research $\&$ engineering atmosphere in the deep graph clustering field, we make efforts to the open-source projects of deep graph clustering. Concretely, we make a deep graph clustering collection and build a unified framework of deep graph clustering methods. We hope these open resources will help the researchers quickly startup, solve critical problems, and apply deep graph clustering methods to a boarder range of applications. Next, we introduce these two GitHub repositories in detail.

\subsubsection{Awesome Deep Graph Clustering}
Awesome Deep Graph Clustering (ADGC) is a collection of state-of-the-art, novel deep graph clustering methods, including papers, codes, and datasets. In addition, this project open-sources some common data processing, clustering, metric calculation, and visualization functions. The project address can be found at GitHub (\url{https://github.com/yueliu1999/Awesome-Deep-Graph-Clustering}). Up to this paper submission time, we have collected about 130 papers and codes and about 20 datasets. Besides, this GitHub repository has gained about 500 stars and 100 forks. We welcome any issues, pull requests, contributions, discussions, etc.

 % and the homepage of Awesome Deep Graph Clustering is demonstrated in Figure \ref{ADGC}
% \begin{figure}[h]
% \centering
% \small
% \begin{minipage}{0.90\linewidth}
% \centerline{\includegraphics[width=1\textwidth]{TPAMI/ADGC.png}}
% \end{minipage}
% \caption{Homepage of Awesome Deep Graph Clustering.}
% \label{ADGC}
% \end{figure}

\subsubsection{A Unified Framework of Deep Graph Clustering}
A-Unified-Framework-for-Deep-Attribute-Graph-Clustering aims to build a unified framework for the deep graph clustering methods. The project address can be found at GitHub (\url{https://github.com/Marigoldwu/A-Unified-Framework-for-Deep-Attribute-Graph-Clustering}). This project redesigns the architecture of the codes of ADGC so that helps researchers conduct experiments easily. The tool classes and functions are defined to simplify the codes and clarify the settings configuration.

% and the homepage of Awesome Deep Graph Clustering is demonstrated in Figure \ref{Unified}
% \begin{figure}[h]
% \centering
% \small
% \begin{minipage}{0.90\linewidth}
% \centerline{\includegraphics[width=1\textwidth]{TPAMI/unified.png}}
% \end{minipage}
% \caption{Homepage of A-Unified-Framework-for-Deep-Attribute-Graph-Clustering.}
% \label{Unified}
% \end{figure}

% Specifically, I redesigned the architecture of the code, so that you can run the open source code easily. I defined some tool classes and functions to simplify the code and make the settings' configuration clear.

\section{Conclusion}
In this work, we make a comprehensive survey about deep graph clustering to help researchers review, summarize, and plan for the future. Concretely, we first demonstrate the general pipeline of deep graph clustering and give a formal definition. Besides, we introduce the critical development and evaluation of deep graph clustering. Secondly, a structured taxonomy is provided from four perspectives, including graph type, network architecture, learning paradigm, and clustering method. Moreover, through extensive experiments, we analyze and summarize the challenges in this domain, such as the problems of graph data quality, stability, scalability, discriminative capability, and unknown cluster number. Also, the potential technical solutions are carefully discussed. Thirdly, we introduce the applications of deep graph clustering in six domains, including social network analysis, recommendation systems, computer vision, natural language processing, bioinformatics, and medical science. More importantly, we contribute two open resources on the GitHub platform. Among them, Awesome Deep Graph Clustering is a comprehensive collection of deep graph clustering, including papers, codes, and datasets. Besides, A-Unified-Framework-for-Deep-Attribute-Graph-Clustering is a unified framework of deep graph clustering. We hope this work becomes a quick guide for the researchers and motivates them to solve significant problems in deep graph clustering.

% introduce the detailed definition and applications of deep graph clustering. 

% Besides, this paper summarizes the important baselines and the taxonomy of deep graph clustering methods based on four criteria. 

% Besides, the collection of deep graph clustering (papers, codes, and datasets) and a unified framework of deep graph clustering are shared on GitHub platform. 

\ifCLASSOPTIONcompsoc
  \section*{Acknowledgments}
\else
  \section*{Acknowledgment}
\fi

% This work was supported by National Key R&D Program of China ()

This work was supported by the National Key R\&D Program of China (project no. 2020AAA0107100, 2021ZD0140406, 2022ZD0115100), the National Natural Science Foundation of China (project no, 62325604, 62276271, U21A20427, 62006237, 61922088, 61906020 and 61773392), the Research Center for Industries of the Future (Project WU2022C043), and the Competitive Research Fund (Project WU2022A009) from the Westlake Center for Synthetic Biology and Integrated Bioengineering. Besides, we thank Benyu Wu (\url{https://marigoldwu.github.io/}) for his significant contributions to A-Unified-Framework-for-Deep-Attribute-Graph-Clustering project.

% We thank all editors and reviewers for their valuable time and constructive suggestions. 

% This work was supported by the National Key R\&D Program of China (project no. 2020AAA0107100) and the National Natural Science Foundation of China (project no. 62006237, 61922088, 61906020 and 61773392).

% Besides, this work was also supported by the National Key R\&D Program of China (Project 2022ZD0115100), the National Natural Science Foundation of China (Project U21A20427), the Research Center for Industries of the Future (Project WU2022C043), and the Competitive Research Fund (Project WU2022A009) from the Westlake Center for Synthetic Biology and Integrated Bioengineering. 

\ifCLASSOPTIONcaptionsoff
  \newpage
\fi

\bibliographystyle{IEEEtran}
\bibliography{ref}

\begin{IEEEbiography}[{\includegraphics[width=1in,height=1.1in,clip,keepaspectratio]{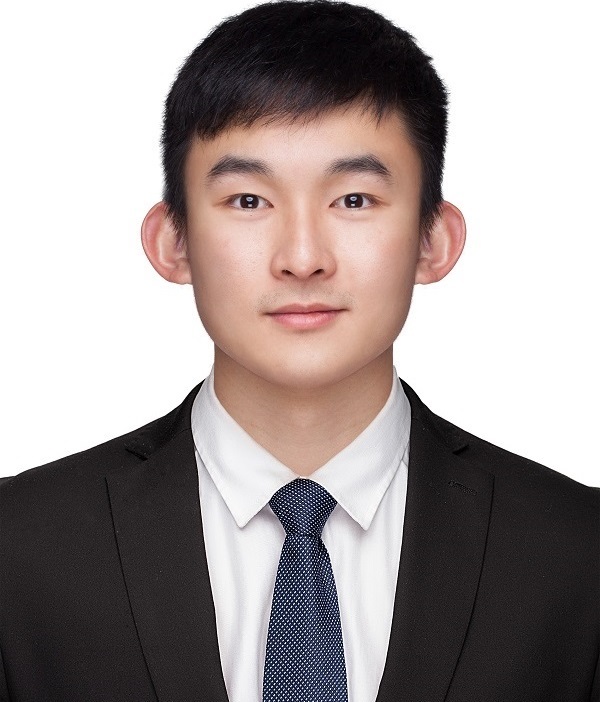}}]
{Yue Liu} is a computer science master's student at National University of Dense Technology (NUDT). His research interests include self-supervised learning, deep graph clustering, and knowledge graphs. He has published 10+ peer-reviewed papers, including ICLR, ICML, AAAI, IJCAI, ACM MM, SIGIR, IEEE T-KDE, and IEEE T-NNLS. More detailed information is listed on his homepage: \url{https://yueliu1999.github.io/}.
\end{IEEEbiography}

\begin{IEEEbiography}[{\includegraphics[width=1in,height=1.10in,clip,keepaspectratio]{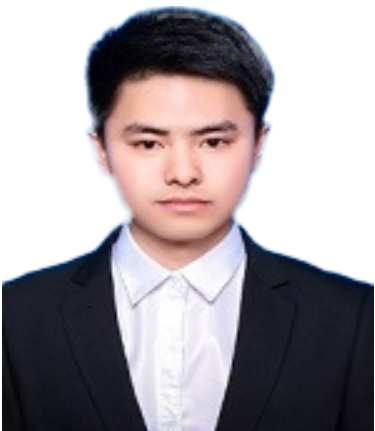}}]
{Jun Xia} received the B. Eng honour degree from Central South University (CSU), Changsha, China in 2020. He is currently pursuing Ph.D. degree at Westlake University and Zhejiang University (ZJU), Hangzhou, China, supervised by Chair Professor Stan Z.Li (IEEE Fellow). His research interests include graph neural networks and AI for scientific discovery. He has published 20+ papers in top-tier conferences such as IEEE T-KDE, ICML, ICLR, CVPR, WWW, ACL, IJCAI, ACM Multimedia, ECML-PKDD, etc.
\end{IEEEbiography}

\begin{IEEEbiography}
[{\includegraphics[width=1in,height=1.15in,clip,keepaspectratio]{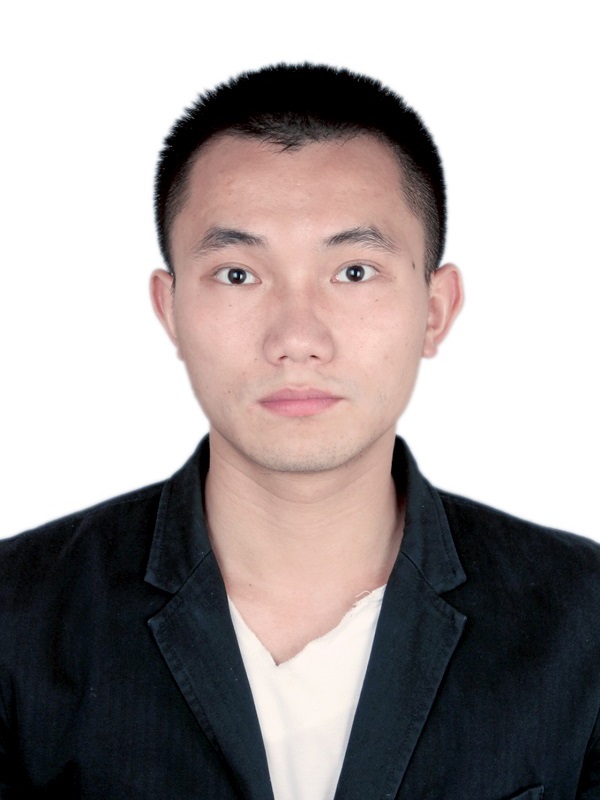}}]{Sihang Zhou} received his PhD degree from School of Computer, National University of Defense Technology (NUDT), China. He is now lecturer at College of Intelligence Science and Technology, NUDT. His current research interests include machine learning and medical image analysis. Dr. Zhou has published 40+ peer-reviewed papers, including IEEE T-IP, IEEE T-NNLS, IEEE T-MI, Information Fusion, Medical Image Analysis, AAAI, MICCAI, etc.
\end{IEEEbiography}

\begin{IEEEbiography}[{\includegraphics[width=1in,height=1.0in,clip,keepaspectratio]{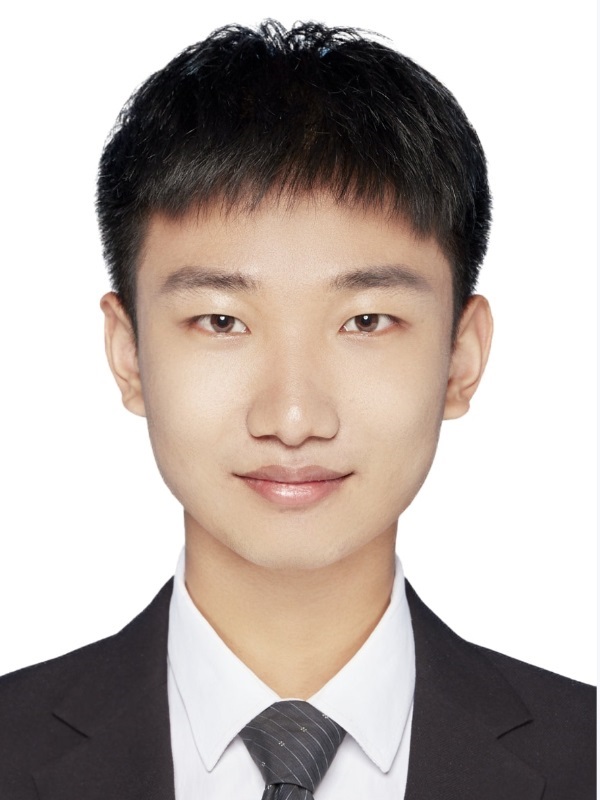}}]
{Xihong Yang} is pursuing his Ph.D. at National University of Defense Technology (NUDT), China. His current research interests include semi-supervised learning, self-supervised learning, and graph neural networks. He has published several papers in top journals and conferences, such as IEEE T-NNLS, IEEE T-KDE, IEEE T-AI, AAAI, ICML, ACM MM, etc.
\end{IEEEbiography}

\begin{IEEEbiography}[{\includegraphics[width=1in,height=1.10in,clip,keepaspectratio]{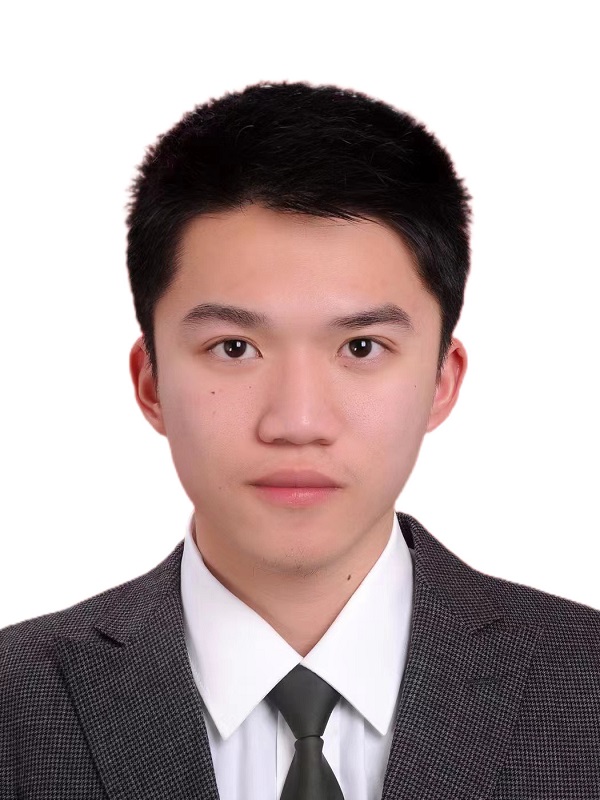}}]
{Ke Liang} is currently pursuing a Ph.D. degree at the National University of Defense Technology (NUDT). Before joining NUDT, he got BSc degree from Beihang University (BUAA) in 2017 and MSc degree from Pennsylvania State University (PSU) in 2021. His research interests include knowledge graphs, graph learning, healthcare AI, and urban computing. He has published several top-tier journal and conference papers, \textit{e.g.,} IEEE T-KDE, IEEE T-NNLS, SIGIR, AAAI, ICML, ACM MM, etc.
\end{IEEEbiography}

\begin{IEEEbiography}[{\includegraphics[width=1in,height=1.10in,clip,keepaspectratio]{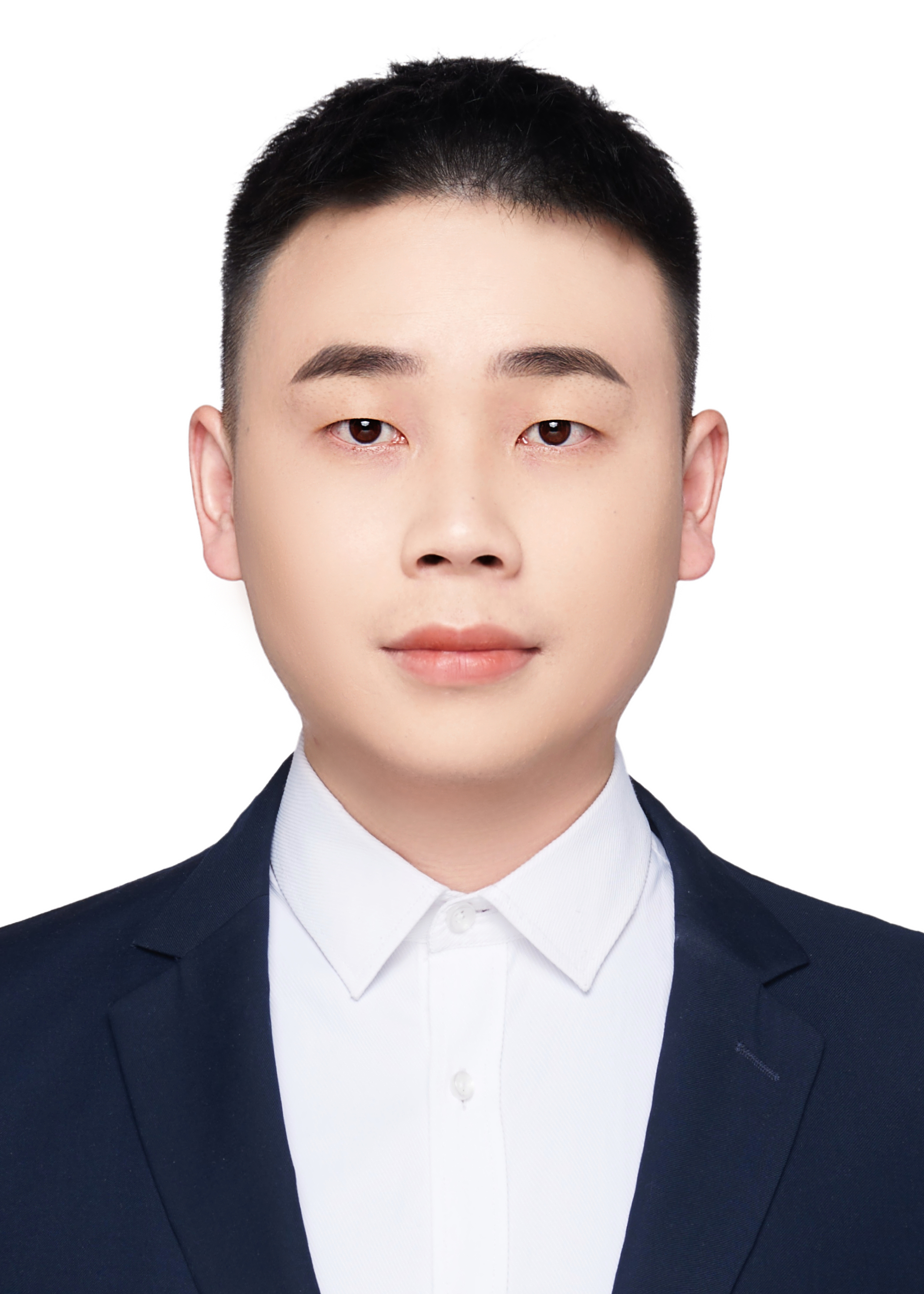}}]
{Chenchen Fan} is a postdoctoral researcher at the Chinese PLA General Hospital. Prior to this, he obtained his PhD from the Institute of Automation, Chinese Academy of Sciences and his bachelor's degree from the Beijing University of Chemical Technology. His research interests include medical artificial intelligence, medical image processing, etc.
\end{IEEEbiography}

\begin{IEEEbiography}[{\includegraphics[width=1in,height=1.10in,clip,keepaspectratio]{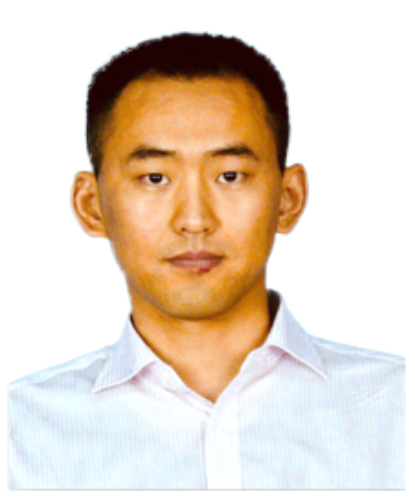}}]{Yan Zhuang} is now a senior engineer at Medical Big Data Research Center, Chinese PLA General Hospital. He has rich experience in hospital informatization and research, expertized in medical big data and artificial intelligence technologies. He has published 20+ papers as the first author and 90+ collaborative papers in the field of medical big data. In addition, He was honored with the Best Paper Award at CIKM 2017.
\end{IEEEbiography}

\begin{IEEEbiography}[{\includegraphics[width=1in,height=1.10in,clip,keepaspectratio]{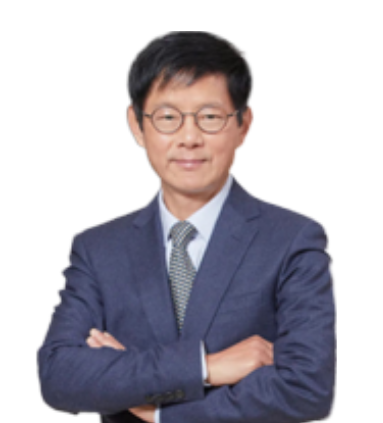}}]{Stan Z. Li} received Ph.D. degree from Surrey University, United Kingdom. He is currently a chair professor and the director of Center for Artificial Intelligence Research and Innovation (CAIRI), School of Engineering, Westlake University. His research interests include pattern recognition, machine learning, image processing, face recognition, biometrics, and intelligent video surveillance. He has published 400+ papers in international journals and conferences and authored and edited eight books. He was an associate editor of the IEEE T-PAMI. He was elevated to IEEE fellow for his contributions to face recognition, pattern recognition, and computer vision, and he is a member of the IEEE Computer Society.
\end{IEEEbiography}

\begin{IEEEbiography}[{\includegraphics[width=1in,height=1.10in,clip,keepaspectratio]{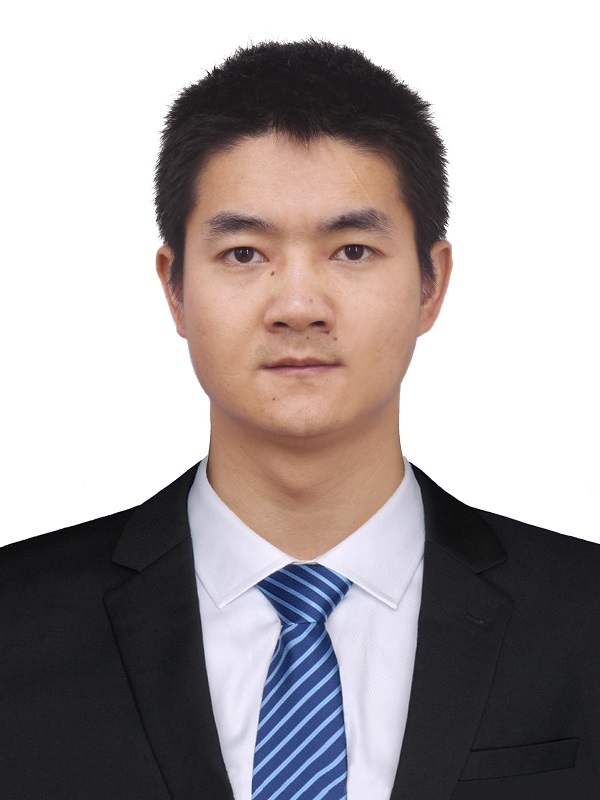}}]{Xinwang Liu} received his PhD degree from National University of Defense Technology (NUDT), China. He is now Professor of School of Computer, NUDT. His current research interests include kernel learning and unsupervised feature learning. Dr. Liu has published 60+ peer-reviewed papers, including those in highly regarded journals and conferences such as IEEE T-PAMI, IEEE T-KDE, IEEE T-IP, IEEE T-NNLS, IEEE T-MM, IEEE T-IFS, ICML, NeurIPS, ICCV, CVPR, AAAI, IJCAI, etc. He serves as the associated editor of T-NNLS, T-CYB and Information Fusion Journal. More information can be found at \url{https://xinwangliu.github.io/}.
\end{IEEEbiography}

\begin{IEEEbiography}[{\includegraphics[width=1in,height=1.10in,clip,keepaspectratio]{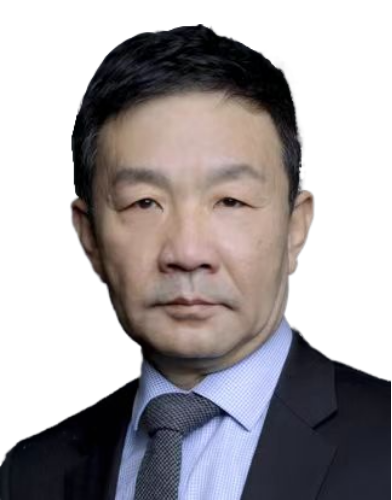}}]{Kunlun He} received his M.D. degree from The 3rd Military Medical University, Chongqing, China, in 1988, and Ph.D. degree in Cardiology from Chinese PLA Medical School, Beijing, China, in 1999. He worked as a postdoctoral research fellow at the Division of circulatory physiology of Columbia University from 1999 to 2003. He is the director and professor of Medical Big Data Research Center, Chinese PLA General Hospital, Beijing, China. His research interests include medical big data and artificial intelligence for cardiovascular disease.
\end{IEEEbiography}

% Kunlun He 

\end{document}